\documentclass{article} 
\usepackage{iclr2027_conference,times}

\usepackage{amsmath,amsfonts,bm}

\def\eqref#1{equation~\ref{#1}}

\def\1{\bm{1}}

\DeclareMathAlphabet{\mathsfit}{\encodingdefault}{\sfdefault}{m}{sl}
\SetMathAlphabet{\mathsfit}{bold}{\encodingdefault}{\sfdefault}{bx}{n}

\usepackage{hyperref}
\usepackage{url}

\usepackage{booktabs}       
\usepackage{amsfonts}       
\usepackage{nicefrac}       
\usepackage{microtype}      
\usepackage{xcolor}         
\usepackage{tikz}
\usetikzlibrary{arrows.meta,positioning,bending}

\usepackage{amsmath}
\usepackage{amssymb}
\usepackage{mathtools}
\usepackage{amsthm}
\let\plainnewtheorem\newtheorem
\usepackage{enumitem}
\usepackage{color}
\usepackage{wrapfig}
\usepackage{multirow}
\usepackage{diagbox}
\usepackage{float}
\usepackage[most]{tcolorbox}
\tcbuselibrary{breakable}
\usepackage{tabularx}

\usepackage{thmtools}
\usepackage{thm-restate}

\theoremstyle{plain}
\plainnewtheorem{theorem}{Theorem}[section]
\plainnewtheorem{proposition}[theorem]{Proposition}
\plainnewtheorem{lemma}[theorem]{Lemma}
\plainnewtheorem{corollary}[theorem]{Corollary}
\theoremstyle{definition}
\plainnewtheorem{definition}[theorem]{Definition}
\plainnewtheorem{assumption}[theorem]{Assumption}
\theoremstyle{remark}
\plainnewtheorem{remark}[theorem]{Remark}

\usepackage[textsize=tiny]{todonotes}
\usepackage[capitalize,noabbrev]{cleveref}

\title{\textsc{ReMAP}: Lossless Residual Neural Lifting for Scalable MAP Inference in Arbitrary-Order Markov Random Fields}

\author{ Yaomin Wang$^{1,2}$, Chaolong Ying$^{1}$, Xiaodong Luo$^{1,2}$, Tianshu Yu$^{1,}$\thanks{the corresponding author}\\
    $^{1}$School of Data Science, The Chinese University of Hong Kong, Shenzhen\\
    $^{2}$Shenzhen Research Institute of Big Data, China \\
    \texttt{\{yaominwang, chaolongying\}@link.cuhk.edu.cn}\\
    \texttt{\{xiaodongluo, yutianshu\}@cuhk.edu.cn}  \\
}

\iclrfinalcopy 
\begin{document}

\maketitle

\begin{abstract}
Maximum a posteriori inference in a finite Markov random field is generally NP-hard. We study an instance-wise optimizer that minimizes the exact multilinear energy under product marginals. The canonical parameterization adds a GNN output to free per-node logits. This residual bypass is a global change of coordinates, so the lifting preserves the infimum, critical points, local extrema (modulo the GNN-parameter fiber), and Hessian inertia of the free-logit problem. Joint gradient flow then induces the preconditioner $I+JJ^{\top}$: the GNN cannot enlarge the feasible set, but it can add dissipation $\|J^{\top}\nabla\ell\|_{2}^{2}$. For a two-layer branch this extra coupling is four-hop local. We also show that every finite critical point of a nonconstant multilinear objective is a saddle, and that a unique MAP gap $\Gamma$ makes conditional-expectation rounding exact once the relaxed gap falls below $\Gamma$; residual gradient flow and joint gradient descent then obey explicit $O(1/t)$ and $O(1/k)$ rates in a small-gap region, with an improved bound under residual alignment. Experiments on synthetic MRFs, UAI 2022 benchmarks, and Physical Cell Identity (PCI) problems evaluate this residual geometry at scale.
\end{abstract}.

\section{Introduction}

Let $x=(x_1,\ldots,x_n)$ be a collection of finite-valued variables and let $E(x)$ be a sum of local factor energies. MAP inference asks for an assignment with minimum energy. This elementary statement covers pairwise MRFs, arbitrary-order factor graphs, weighted constraint networks, and many application-specific discrete objectives in computer vision, language processing, communications, and network optimization~\citep{WANG20131610,9112323,article_nlp_mrf_sur,ammar2014conditionalrandomfieldautoencoders,LIN2020431,Wu_Lian_Xu_Wu_Chen_2020,transport}. It is also NP-hard in general~\citep{koller2009probabilistic}. Approximate message-passing methods such as LBP and TRBP are efficient but often degrade on hard dense or high-order problems~\citep{JMLR:v6:ihler05a,NIPS2011_cee63112,1522634,wainwright2005map}. Exact branch-and-bound solvers can certify optimality, and Toulbar2~\citep{de2023toulbar2} is among the strongest general-purpose MAP solvers, having won all MPE and MMAP categories in the UAI 2022 Inference Competition. Even so, finding high-quality incumbents within a practical time budget remains challenging on large or high-order instances.

Neural approaches have shown promise for combinatorial optimization and probabilistic inference, but their use in MRF MAP inference has mostly been limited to small pairwise graphs, supervised imitation of classical solvers, or amortized inference across training distributions~\citep{pmlr-v130-garcia-satorras21a,NEURIPS2020_07217414,yoon2019inferenceprobabilisticgraphicalmodels,pmlr-v180-cui22a}. These settings do not fully address large arbitrary-order MRFs with heterogeneous label cardinalities. Related instance-wise or self-supervised neural methods, including PI-GNN~\citep{Schuetz_2022}, SSMP~\citep{Arya_Rahman_Gogate_2024}, and SINE~\citep{arya2025sine}, likewise do not make ``no labels'' itself a standalone contribution. The central question is whether a graph-structured lifting supplies a useful optimization geometry without sacrificing the free-logit problem.

\textsc{ReMAP} explores this trade-off. For one input instance, a GNN and a free per-node residual produce categorical logits. The expected MRF energy under the resulting product distribution is the standard multilinear relaxation~\citep{lehuu2018continuous}; it is differentiable, so all variables can be optimized on a GPU. A discrete incumbent is extracted during the run and all neural parameters are discarded afterward. Thus, ``training'' is simply the numerical solution of one lifted optimization problem. The starting observation is concrete: if a GNN branch is added residually to free logits, $z=u+\rho_v(G)$, the neural map becomes a global coordinate lifting of the original problem. This gives landscape equivalence, modulo the GNN-parameter fiber, and identifies the only possible optimization benefit---an adaptive positive-semidefinite preconditioner induced by the GNN Jacobian. It also makes the necessary experimental comparison unambiguous: residual \textsc{ReMAP} versus the identical free-logit optimizer and versus the GNN branch alone.

The theoretical analysis specializes locality statements to a two-layer message-passing branch. The lossless-lifting equivalence itself holds for any continuously differentiable GNN branch. In the experiments we retain the original multi-layer settings: a $5$-layer GNN on synthetic and PCI instances and an $8$-layer GNN on UAI instances.

This paper makes four contributions.
\begin{itemize}[leftmargin=*]
    \item We introduce a residual GNN lifting $z=u+\rho_v(G)$ whose parameter-to-logit map is globally equivalent to a change of coordinates. It is lossless in value and preserves the critical-point and local-extremum structure of free logits, with additional flat directions along the GNN-parameter fiber.
    \item We show that joint gradient flow induces the adaptive metric $I+JJ^{\top}$. We quantify its exact additional dissipation, derive an $O(1/T)$ logit-stationarity guarantee, and show that the node-embedding metric of a two-layer GNN is four-hop sparse. Under residual alignment $\|J_v^{\top}q\|_{2}^{2}\ge\beta\|q\|_{2}^{2}$, the stationarity bound improves by the factor $(1+\beta)^{-1}$.
    \item We sharpen the underlying MAP theory: a nonconstant multilinear objective has no finite interior local extrema; a unique MAP gap gives exact conditional-rounding and argmax recovery thresholds; and small-gap gradient flow and joint gradient descent converge in objective and product-distribution distance at explicit $O(1/t)$ and $O(1/k)$ rates, with the constant $c_0$ replaced by $(1+\beta)c_0$ under residual alignment.
    \item We evaluate residual \textsc{ReMAP} on synthetic pairwise and high-order MRFs, UAI 2022 inference benchmarks, and real-world Physical Cell Identity (PCI) problems, and isolate the residual, GNN-only, and free-logit parameterizations on the five real PCI cases.
\end{itemize}

\textbf{Related work.} A detailed comparison with Le-Huu--Paragios, PI-GNN, SSMP, SINE, and learned message-passing methods is given in Appendix~\ref{app:related}.

\section{Preliminaries}\label{sec:pre}

\textbf{Markov Random Field}.~~
An MRF is defined over an undirected graph $\mathcal{G}= (\mathcal{V}, \mathcal{C})$, where $\mathcal{V}$ represents the index set of random variables and $\mathcal{C}\subseteq 2^\mathcal{V}$ is the clique set representing the (high-order) dependencies among random variables. Throughout this paper, we associate each node index $i$ with a random variable $x_i\in\mathcal{X}$, where $\mathcal{X}$ is a finite alphabet. Thus, given graph $\mathcal{G}$, the joint probability of a configuration of $X=\{x_i\}_{i\in\mathcal{V}}$ can be expressed as Eq.~\ref{eq:joint_dis},
\begin{equation}
    \begin{aligned}
    \mathbb{P}(X) = \frac{1}{Z} \exp(-E(X)) 
                  = \frac{1}{Z} \exp\left(-\sum_{i\in\mathcal{V}} \theta_i(x_i) - \sum_{C_k\in \mathcal{C}}  \theta_{C_k}(\{x_l | \forall x_l \in C_k\})\right)
\label{eq:joint_dis}
\end{aligned}
\end{equation}
where $Z$ is the partition function, $\theta_i(\cdot)$ denotes the unary energy functions, $\theta_{C}(\cdot)$ represent the clique energy functions. In this sense, MRF provides a compact representation of probability by introducing conditional dependencies:
\begin{equation}\label{eq:cond_dependency}
     \mathbb{P}(x_i | X\backslash \{x_i\}) = \mathbb{P}(x_i | \{x_j\}~~\text{for}~~i, j \in C_k~~\text{for}~~C_k \in \mathcal{C}).
\end{equation}
In this paper, we consider the MAP estimate of Eq.~\ref{eq:joint_dis}. MAP inference asks for an assignment $x^\star \in \arg\min_x E(x)$, with optimal value $E^\star=\min_x E(x)$. One can consult \citep{koller2009probabilistic} for more details.

\textbf{Graph Neural Networks}.~~
GNNs represent a distinct class of neural network architectures specifically engineered to process graph-structured data~\citep{kipf2017semi,hamilton2017inductive,xu2019powerful, velickovic2018graph}.  
In general, when addressing a problem involving a graph $\mathcal{G}=(\mathcal{V},\mathcal{E})$, where $\mathcal{E}$ is the edge set, GNNs utilize both the graph $\mathcal{G}$ and the initial node representations $\{h_i^{(0)} \in \mathbb{R}^{d} | \forall i \in \mathcal{V} \}$ as inputs, where $d$ is the dimension of initial features. 
Assuming the total number of GNN layers to be $K$, at the $k$-th layer the graph convolutions typically read:
\begin{equation}
h_i^{(k)}=\sigma \left(W_k \cdot \operatorname{AGGREGATE}^{(k)}\left(\left\{h_j^{(k-1)}: j \in \mathcal{N}(i) \cup \{i\}\right\}\right) \right)
\end{equation}
where $\operatorname{AGGREGATE}^{(k)}$ is defined by the specific model, $W_k$ is a trainable weight matrix, $\mathcal{N}(i)$ is the neighborhood of node $i$, and $\sigma$ is a non-linear activation function, e.g., $\mathrm{ReLU}$.

\begin{figure*}[tb]
  \centering
  \begin{tikzpicture}[
      font=\small,
      box/.style={
        draw=black,
        rounded corners=2.5pt,
        align=center,
        inner sep=5pt,
        minimum height=1.45cm,
        text width=2.85cm,
        line width=0.6pt,
      },
      mainbox/.style={box, fill=white},
      decbox/.style={box, fill=green!12},
      arr/.style={-{Latex[length=2.1mm]}, thick, draw=blue!60!black},
    ]
    \node[mainbox] (mrf) {MRF factors\\$\{\theta_C\}_{C\in\mathcal{C}}$};
    \node[mainbox, right=0.42cm of mrf] (lift)
      {free logits $u$ plus\\GNN residual $\rho_v$\\$z=u+\rho_v(G)$};
    \node[mainbox, right=0.42cm of lift] (soft)
      {masked softmax\\$p(w)\in\mathcal{P}$};
    \node[mainbox, right=0.42cm of soft] (loss)
      {multilinear loss\\$F(p)$};
    \node[decbox, below=0.58cm of soft] (ce)
      {conditional\\rounding};
    \node[decbox, right=0.42cm of ce] (inc)
      {best discrete\\incumbent $x^{\mathrm{best}}$};

    \draw[arr] (mrf) -- (lift);
    \draw[arr] (lift) -- (soft);
    \draw[arr] (soft) -- (loss);
    \draw[arr] (soft) -- (ce);
    \draw[arr] (ce) -- (inc);
    \draw[arr]
      (soft.north) to[out=90, in=90, looseness=0.7]
      node[midway, above, font=\scriptsize, align=center, inner sep=0.5pt]
      {gradient update\\for this instance}
      (lift.north);
  \end{tikzpicture}
  \caption{Residual \textsc{ReMAP} is an optimizer for one instance. The factor energy supplies the loss; the free-logit bypass preserves every search direction; the GNN supplies an adaptive graph-structured geometry. Conditional-expectation rounding converts every relaxed iterate into a no-worse discrete incumbent.}
  \label{fig:flow}
\vspace{-10pt}
\end{figure*}

\section{Methodology}\label{sec:method}


An overview of residual \textsc{ReMAP} is in Fig.~\ref{fig:flow}. Given one MRF instance, we expand each factor into pairwise GNN edges, optimize free per-node logits together with a GNN residual, evaluate the exact multilinear energy under the resulting product distribution, and extract a discrete incumbent by conditional-expectation rounding. All parameters are reset for the next instance. The factor tables---not the clique expansion---define the loss. The residual bypass $z=u+\rho_v(G)$ is what makes the lifting globally lossless: the GNN supplies an adaptive graph-structured geometry, while free logits preserve every search direction of the original continuous problem.

\subsection{Preprocessing} \label{sec:preprocessing}

\textbf{Topology construction for GNNs.} In an MRF instance, the high-order graph structure consists of nodes and cliques, diverging from typical GNNs allowing only pairwise edges (2nd-order). To facilitate the power of GNNs, we need to convert the high-order factor structure into a pairwise graph.

By the very definition of a clique, any two nodes that appear within the same clique are directly related. Thus, for any two nodes $i,j\in C_k$ in a clique $C_k$, we add a pairwise edge $(i,j)$ to its GNN-oriented graph. An edge may appear in multiple cliques; we add each edge only once. This expanded pairwise graph is the support of the GNN branch $\rho_v(G)$ in Fig.~\ref{fig:flow}; the factor tables, not the extra edges, define $F(p)$.

\textbf{Initial features for GNNs.} 
As there are no initial features associated with MRF instances,
we initialize feature vectors to GNNs \emph{randomly} with a predefined dimension $d$. Detailed information on how we will handle these artificial features to ensure they effectively capture the underlying information of the problem will be provided in Section~\ref{sec:GNN_lifting}.

\textbf{Vectorizing the energy function.}
The transformed energy function $E(X)$ will serve as the loss function guiding the training of the neural networks. In Section~\ref{sec:loss_func}, we will detail the transformation process and discuss how to effectively utilize it.  
Note the values of these functions can be pre-evaluated and repeatedly used during the training process. 
Therefore, we employ a look-up table to memorize all function values with discrete inputs.
For unary energies, we denote the vectorized unary energy of variable $x_i$ as $\phi(x_i)$, 
where the $n$-th element corresponds to $\theta_i(x_i = n)$. Similarly, we represent the clique energy for clique $C_k$ using the tensor $\psi(\{x_l | \forall x_l \in C_k\})$. This tensor can be derived using the same conceptual framework; for instance, the element $\psi(x_i, x_j, x_k)$ at position $(0, 2, 4)$ corresponds to the value of $\theta_{\{i,j,k\}}(x_i=0, x_j=2, x_k=4)$.

\textbf{Padding node embeddings \& energy terms and Masking.} 
GNNs typically require all node embeddings to be of the same dimension, meaning that the embeddings $h^{(K)}$ at the $K$-th layer must share the same size. However, in general MRFs, the variables often exhibit different numbers of states.
While traditional belief-propagation-based methods can easily manage such variability, adapting GNNs to handle these discrepancies is less straightforward. 

To address this mismatch, we pad node embeddings and unary/pairwise (or clique) energy tables to a common cardinality $|\mathcal{X}|$ so that batched GNN tensors have a uniform last dimension. Concretely, we introduce dummy states for nodes with $s_i<|\mathcal{X}|$ and fill the extra table slices with an arbitrary placeholder (implementation: the original table maximum). The schematic diagram is in Appendix~\ref{sec:padding}. Padding is only a layout device: invalid logits are set to $-\infty$ before softmax, so dummy labels have probability exactly zero. Once this mask is applied, the placeholder values stored in padded energy regions cannot affect $F(p)$ or the decoder.

\begin{remark}
If invalid logits were left finite and padded energies were filled with large constants, those dummy labels could still leak into the multilinear contraction. Masking with $-\infty$ before softmax removes this leakage: padded table entries are multiplied by zero probability and are irrelevant to both the loss and the decoder.
\end{remark}

\subsection{Residual GNN lifting} \label{sec:GNN_lifting}

Let $D=\sum_{i}s_i$ be the number of valid logits and write $z=(z_i)_{i}\in\mathbb{R}^{D}$. The canonical parameterization uses free logits $u\in\mathbb{R}^{D}$ and a GNN branch $\rho_v(G)\in\mathbb{R}^{D}$:
\begin{equation}
    z(u,v)=u+\rho_v(G),\qquad
    p_i(z;\tau)=\mathrm{softmax}(z_i/\tau),\qquad
    L_{\tau}(u,v)=\ell_{\tau}(z(u,v))=F(p(z;\tau)).
    \label{eq:residual}
\end{equation}
All parameters $(u,v)$ are reset and optimized for one instance. A fixed $\tau=1$ is canonical; annealing is optional and lies outside the fixed-temperature theorems. Adding a constant to all logits of one node leaves $p_i$ unchanged; we retain these harmless gauge directions.

The graph branch maps every factor to a clique, starts from trainable node embeddings $h_i^{(0)}\in\mathbb{R}^{d_l}$, applies message-passing layers, and decodes node states into the valid coordinates of $\rho_v(G)$. The intuition is the same as in loopy belief propagation: incoming messages aggregate information from subgraphs that a node cannot access directly. Node $i$ updates its beliefs by combining unary information with neighborhood messages,
\begin{equation}
    p^{\textbf{posterior}}(x_i | X\backslash \{x_i\}) = \exp(-\phi(x_i)) \prod \limits_{(i,j) \in \mathcal{E}} \sum \limits_{x_j} M_{ji},
    \label{eq:BP}
\end{equation}
which is structurally analogous to GNN message passing.

Graph convolutions should treat adjacent nodes equally, consistent with MRF neighborhood aggregation. Typical GNNs are summarized in Table~\ref{table:GNNs} in Appendix~\ref{sec:GNNs}. We use the GraphSAGE aggregator~\citep{hamilton2017inductive} as the backbone; GCN~\citep{kipf2017semi} and GAT~\citep{velickovic2018graph} are compared in Fig.~\ref{fig:GNN_backbone_curve}. We further use Jumping Knowledge~\citep{xu2018representationlearninggraphsjumping} so that local and longer-range features can be combined. For Hessian statements we assume a twice continuously differentiable activation; first-order results require differentiability only along the trajectory.

The theoretical locality result in Section~\ref{sec:theory} is stated for a two-layer branch: if $h_i^{(0)}=e_i$ and $k\in\{1,2\}$, a GraphSAGE-style update is given by
\begin{equation}
    m_i^{(k)}=\frac{1}{|\mathcal{N}(i)|}\sum_{j\in\mathcal{N}(i)}h_j^{(k-1)},\qquad
    h_i^{(k)}=\sigma\bigl(A_k h_i^{(k-1)}+B_k m_i^{(k)}+b_k\bigr)
    \label{eq:twolayer}
\end{equation}
For isolated nodes, the neighbor aggregate is taken to be zero.
This update yields a four-hop node-embedding metric. A Jumping-Knowledge decoder then maps $[h_i^{(0)},h_i^{(1)},h_i^{(2)}]$ to $\rho_{v,i}(G)$. In experiments we keep the original deeper stacks ($K=5$ on synthetic and PCI instances, $K=8$ on UAI instances). The lossless-lifting equivalence of Theorem~\ref{thm:lossless} does not depend on width, depth, or expressivity; only the four-hop sparsity of Proposition~\ref{prop:fourhop} is two-layer specific.

The free logits, embeddings, shared matrices, and decoder are optimized jointly. Invalid logits induced by label-cardinality padding are set to $-\infty$ before softmax, so their probabilities are exactly zero and padded energy entries cannot affect $F(p)$.

\subsection{Energy minimization and incumbent extraction} \label{sec:loss_func}
Let $\Delta_i=\{p_i\in\mathbb{R}_{+}^{s_i}:\mathbf{1}^{\top}p_i=1\}$ and $\mathcal{P}=\prod_i\Delta_i$. The multilinear objective~\citep{lehuu2018continuous} is
\begin{equation}
    F(p)=\sum_{C\in\mathcal{C}}\sum_{x_C}\theta_C(x_C)\prod_{i\in C}p_i(x_i).
    \label{eq:multilinear}
\end{equation}
Equivalently, writing $v_i$ for a one-hot encoding of $x_i$ and $P_k=\otimes_{i\in C_k}p_i$,
\begin{equation}
    E(\{v_i | i \in \mathcal{V}\}) = \underbrace{\sum \limits_{i \in \mathcal{V}} \langle v_i, \phi(x_i) \rangle}_{\text{Unary Term}} +
                \underbrace{\sum \limits_{C_k \in \mathcal{C}}  \langle \psi(C_k), V_k \rangle }_{\text{Clique Term}}
    \label{eq:loss_func_v}
\end{equation}
\begin{equation}
    F(p) = \underbrace{\sum \limits_{i \in \mathcal{V}} \langle p_i, \phi(x_i) \rangle}_{\text{Unary Term}} +
                 \underbrace{\sum \limits_{C_k \in \mathcal{C}}  \langle \psi(C_k), P_k \rangle }_{\text{Clique Term}}.
    \label{eq:loss_func_p}
\end{equation}
This is the expected energy under the product distribution $\prod_i p_i$; it is affine in each block $p_i$ but not jointly convex. Residual \textsc{ReMAP} optimizes $L_{\tau}(u,v)=F(p(z(u,v);\tau))$ from Eq.~\ref{eq:residual}. Exactness belongs to $F$ on $\mathcal{P}$, not to the GNN.

At every checkpoint we decode with conditional-expectation rounding and retain the lowest-energy incumbent. This decoder returns a discrete assignment $x_{\mathrm{CE}}$ with $E(x_{\mathrm{CE}})\le F(p)$ for every $p\in\mathcal{P}$. Node-wise argmax is a cheaper optional decoder and receives a separate confidence bound in Section~\ref{sec:theory}; all reported experiments use conditional-expectation rounding. We use Adam~\citep{kingma2014adam} as a practical optimizer; the dynamics below analyze exact gradient flow and gradient descent. Annealing, when used, is a heuristic outside the fixed-temperature theorems.

\section{Theory: Lossless Lifting and Graph-Structured Optimization}\label{sec:theory}
Proofs of all formal statements are deferred to Appendix~\ref{app:proofs}. Write $E_{\star}=\min_{x\in\mathcal{X}}E(x)$ and, for a variable $i$, the conditional cost
\begin{equation}
    g_i(a;p_{-i})=\mathbb{E}_{X_{-i}\sim\prod_{j\neq i}p_j}[E(a,X_{-i})],\qquad F(p)=\langle p_i,g_i(p_{-i})\rangle.
    \label{eq:condcost}
\end{equation}

\begin{restatable}{theorem}{thmid}[Expectation identity, exact value, and minimizers]
\label{thm:identity}
Let $X_i\sim p_i$ independently. For every $p\in\mathcal{P}$,
\begin{equation}
    F(p)=\mathbb{E}_{X\sim\prod_i p_i}[E(X)]\ge E_{\star}.
\end{equation}
Moreover, $\min_{p\in\mathcal{P}}F(p)=E_{\star}$. Equality $F(p)=E_{\star}$ holds if and only if every assignment in $\mathrm{supp}(\prod_i p_i)$ is globally optimal. In particular, if the MAP assignment is unique, the unique relaxed minimizer is its one-hot encoding.
\end{restatable}

The theorem separates exactness of the objective on $\mathcal{P}$ from expressivity of its parameterization.

\begin{restatable}{corollary}{corce}[Conditional-expectation rounding]
\label{cor:ce}
For every $p\in\mathcal{P}$, a deterministic assignment $x_{\mathrm{CE}}$ can be constructed sequentially such that $E(x_{\mathrm{CE}})\le F(p)$. At step $i$, set a label minimizing the expected energy when that label and all earlier choices are fixed while unrounded variables retain their current marginals. The guarantee holds for every variable order and arbitrary finite factors.
\end{restatable}

This is the method of conditional expectations and is closely related to block rounding in L\^{e}-Huu and Paragios~\citep{lehuu2018continuous}. For cases where argmax is retained for speed, define $\mathrm{osc}(\theta_C)=\max_{x_C}\theta_C(x_C)-\min_{x_C}\theta_C(x_C)$.

\begin{restatable}{proposition}{propargmax}[Argmax-rounding deviation]
\label{prop:argmax}
Let $\hat{x}_i\in\arg\max_a p_i(a)$. Then
\begin{equation}
    \bigl|E(\hat{x})-F(p)\bigr|
    \le \sum_{C\in\mathcal{C}}\mathrm{osc}(\theta_C)\Bigl(1-\prod_{i\in C}p_i(\hat{x}_i)\Bigr)
    \le \sum_{C\in\mathcal{C}}\mathrm{osc}(\theta_C)\sum_{i\in C}\bigl(1-p_i(\hat{x}_i)\bigr).
\end{equation}
Thus the argmax gap is controlled by both distributional concentration and factor ranges.
\end{restatable}

\paragraph{What finite softmax can represent.}
Finite logits parameterize the relative interior of each simplex and cannot produce a one-hot vector. The closed product of simplices attains $\min_{p\in\mathcal{P}}F(p)=E_{\star}$ at vertices (Theorem~\ref{thm:identity}); finite softmax generally only attains an infimum, approached as $\|z\|\to\infty$.

\begin{restatable}{theorem}{thmsoftmax}[Finite-softmax infimum and a quantitative approximation]
\label{thm:softmax}
Let $z=(z_i)_i$ be free finite logits and $p_i=\mathrm{softmax}(z_i/\tau)$ for fixed $\tau>0$. Then $\inf_z F(p(z))=E_{\star}$. The infimum is attained at finite $z$ if and only if every assignment is globally optimal. More quantitatively, let $x^{\star}$ be optimal, let $R_E=\max_x E(x)-E_{\star}$, and set $z_i(x_i^{\star})=m$, $z_i(a)=0$ for $a\neq x_i^{\star}$. Then
\begin{equation}
    0\le F(p(z))-E_{\star}\le R_E\sum_{i=1}^{n}(s_i-1)e^{-m/\tau}.
\end{equation}
\end{restatable}

The result explains why a convergence theorem for the parameter iterates is the wrong default target: even in the free-logit model, a loss-minimizing sequence may require $\|z\|\to\infty$.

\paragraph{A lossless residual lifting.}
Let $v\in\mathcal{V}\subseteq\mathbb{R}^{m}$, with $\mathcal{V}$ open, parameterize any continuously differentiable GNN output $\rho_v(G)\in\mathbb{R}^{D}$. The next result holds regardless of width, depth, activation, factor order, or expressivity; its Hessian statement additionally assumes $\rho$ is twice continuously differentiable.

\begin{restatable}{theorem}{thmlossless}[Global lossless-lifting equivalence]
\label{thm:lossless}
Define $T(u,v)=(u+\rho_v(G),v)$. Then $T$ is a global diffeomorphism with inverse $T^{-1}(z,v)=(z-\rho_v(G),v)$, and $L_{\tau}\circ T^{-1}(z,v)=\ell_{\tau}(z)$. Consequently:
(i) $\inf_{u,v}L_{\tau}(u,v)=\inf_z\ell_{\tau}(z)=E_{\star}$;
(ii) $(u,v)$ is critical, a local minimum, or a local maximum of $L_{\tau}$ if and only if $z=u+\rho_v(G)$ has the corresponding property for $\ell_{\tau}$;
(iii) every sublevel set of $L_{\tau}$ is diffeomorphic to the corresponding free-logit sublevel set times $\mathcal{V}$;
(iv) at a critical point, the Hessian of $L_{\tau}$ has the same positive and negative inertia as $\nabla^2\ell_{\tau}(z)$, with at least $m$ additional zero directions.
\end{restatable}

In the transformed coordinates $(z,v)$, the lifted objective equals $\ell_{\tau}(z)$ and is constant along the $v$-fiber. Hence $(u,v)$ is a critical point or local extremum of $L_{\tau}$ if and only if $z=u+\rho_v(G)$ is a critical point or local extremum of $\ell_{\tau}$; the additional zero Hessian directions in (iv) are these fiber directions. Theorem~\ref{thm:lossless} is a general landscape-equivalence statement. In the nonconstant MAP case, Theorem~\ref{thm:saddle} further implies that finite stationary points are saddles rather than local extrema. Any optimization benefit must come from the metric, not from a larger set of logits.

\paragraph{Identity-plus-GNN preconditioning.}
Write $J_v=D_v\rho_v(G)$ and $q=\nabla\ell_{\tau}(z)$. Residual alignment of strength $\beta\ge 0$ means
\begin{equation}
    \|J_v^{\top}q\|_{2}^{2}\ge\beta\|q\|_{2}^{2}.
    \label{eq:align}
\end{equation}
When $q\neq 0$, the diagnostic ratio is $\beta_t=\|J_v^{\top}q\|_{2}^{2}/\|q\|_{2}^{2}$. This is a squared-norm ratio: the joint gradient satisfies
\begin{equation}
    \|\nabla_{u,v}L_{\tau}\|_{2}^{2}=\|q\|_{2}^{2}+\|J_v^{\top}q\|_{2}^{2}=q^{\top}(I+J_v J_v^{\top})q\ge(1+\beta)\|q\|_{2}^{2},
    \label{eq:gradamp}
\end{equation}
so any stationarity or descent bound written in terms of $\|q\|_{2}^{2}$ improves by the factor $(1+\beta)^{-1}$. Alignment is instance-, parameter-, and trajectory-dependent; the architecture does not force $\beta>0$, and $\beta=0$ recovers the free-logit bound.

\begin{restatable}{theorem}{thmdyn}[Induced logit dynamics and dissipation]
\label{thm:dyn}
Along $\dot u=-\nabla_u L_{\tau}$ and $\dot v=-\nabla_v L_{\tau}$, the logits obey $\dot z=-M_v q$ with $M_v=I_D+J_v J_v^{\top}\succeq I_D$, and
\begin{equation}
    \frac{d}{dt}L_{\tau}=-q^{\top}M_v q=-\|q\|_{2}^{2}-\|J_v^{\top}q\|_{2}^{2}.
    \label{eq:dissipation}
\end{equation}
For every $T>0$, $\min_{0\le t\le T}\|q(t)\|_{2}^{2}\le (L_{\tau}(0)-E_{\star})/T$. If $\|J_v^{\top}q\|_{2}^{2}\ge\beta\|q\|_{2}^{2}$ throughout the interval, the right-hand side improves by the factor $(1+\beta)^{-1}$.
\end{restatable}

Equation~\ref{eq:dissipation} is a pointwise comparison at the same logit state $z$, not a claim that two different trajectories reach the global solution in the same time. The extra graph contribution is the squared alignment ratio $\beta_t$ in Eq.~\ref{eq:align}. It is not identified by comparing terminal energies of residual, GNN-only, and free parameterizations, which follow different trajectories.

\begin{restatable}{corollary}{cordisc}[Discrete-time stationarity]
\label{cor:disc}
Suppose $L_{\tau}$ has $H$-Lipschitz gradient on a convex region containing each segment between $K$ consecutive joint gradient-descent iterates, and use $0<\eta\le H^{-1}$. Then
$\min_{0\le k<K}\|\nabla\ell_{\tau}(z_k)\|_{2}^{2}\le 2(L_{\tau}(u_0,v_0)-E_{\star})/(\eta K)$.
If $\|J_{v_k}^{\top}q_k\|_{2}^{2}\ge\beta\|q_k\|_{2}^{2}$ at all iterates, the right-hand side improves by $(1+\beta)^{-1}$.
\end{restatable}

\begin{restatable}{proposition}{propfourhop}[Four-hop locality of a two-layer branch]
\label{prop:fourhop}
Partition $v=(e,\omega)$ into node embeddings and shared GNN weights. If two message-passing layers are used, then $M_v=I_D+J_e J_e^{\top}+J_{\omega}J_{\omega}^{\top}$, and the node block $(J_e J_e^{\top})_{ik}$ is zero whenever $\mathrm{dist}_G(i,k)>4$. The embedding metric is therefore graph local; the shared-weight term can add a global structured coupling.
\end{restatable}

\paragraph{Finite critical points are saddles.}
\begin{restatable}{theorem}{thmsaddle}[No finite-logit local extrema]
\label{thm:saddle}
If $E$ is not constant on $\mathcal{X}$, neither $\ell_{\tau}$ nor residual $L_{\tau}$ has a finite local minimum or maximum. A finite logit $z$ is critical exactly when every $g_i(\cdot;p_{-i})$ is constant. Every such point is a topological saddle: every neighborhood contains both lower- and higher-valued points. If its Hessian is nonzero, the Hessian has at least one positive and one negative eigenvalue. Residual lifting preserves these signs and adds flat fiber directions.
\end{restatable}

The absence of finite local minima does not imply global convergence: the infimum normally lies at diverging logits, saddle manifolds may be degenerate, and discrete coordinate-wise minima reappear on the boundary of $\mathcal{P}$.

\paragraph{MAP recovery and local convergence.}
Assume in this subsection that $x^{\star}$ is the unique MAP assignment. Define
\begin{equation}
    \Gamma=\min_{x\neq x^{\star}}E(x)-E_{\star},\qquad
    R_E=\max_x E(x)-E_{\star},\qquad
    \kappa=\min_i\min_{a\neq x_i^{\star}}E(a,x_{-i}^{\star})-E_{\star}.
\end{equation}

\begin{restatable}{theorem}{thmgap}[Gap thresholds for exact decoding]
\label{thm:gap}
Let $\Delta(p)=F(p)-E_{\star}$. Then $\Pr_{X\sim\prod_i p_i}(X\neq x^{\star})\le\Delta(p)/\Gamma$. If $\Delta(p)<\Gamma$, conditional-expectation rounding returns $x^{\star}$ for every variable order. If $\Delta(p)<\Gamma/2$, node-wise argmax also returns $x^{\star}$ uniquely.
\end{restatable}

The first threshold is useful even though every finite-softmax marginal remains strictly fractional: once the relaxed objective falls below the second-best discrete energy, the safe decoder is already exact.

\begin{restatable}{theorem}{thmflow}[Small-gap convergence of residual gradient flow]
\label{thm:flow}
Let $\varepsilon_0=\min\{\Gamma/2,\kappa\Gamma/(4n R_E)\}$ and $c_0=\kappa^2/(16\tau^2 n R_E^2)$. If $\Delta(0)\le\varepsilon_0$, then residual gradient flow remains in this region and $\Delta(t)\le\Delta(0)/(1+c_0\Delta(0)t)$. If $\|J_v^{\top}q\|_{2}^{2}\ge\beta\|q\|_{2}^{2}$ along the trajectory, replace $c_0$ by $(1+\beta)c_0$. Conditional-expectation decoding equals $x^{\star}$ throughout this region, and the induced product $Q_t=\prod_i p_i(t)$ satisfies $\|Q_t-\delta_{x^{\star}}\|_{\mathrm{TV}}\le \Delta(0)/(\Gamma(1+c_0\Delta(0)t))$.
\end{restatable}

The rate is local and has conservative constants. It does not assume convexity or bounded logits.

\begin{restatable}{corollary}{corgd}[Small-gap rate for joint gradient descent]
\label{cor:gd}
Under the assumptions of Theorem~\ref{thm:flow}, let joint gradient descent update $(u,v)$ with step size $0<\eta\le H^{-1}$. Suppose $L_{\tau}$ has $H$-Lipschitz gradient on every iterate segment. If $\Delta_0\le\varepsilon_0$, then all iterates remain in the same small-gap region and $\Delta_k\le\Delta_0/(1+(\eta c_0/2)\Delta_0 k)$. If $\|J_{v_k}^{\top}q_k\|_{2}^{2}\ge\beta\|q_k\|_{2}^{2}$ at every nonstationary iterate, replace $c_0$ by $(1+\beta)c_0$. The total-variation and exact conditional-decoding conclusions of Theorem~\ref{thm:flow} hold with the discrete rate in place of the continuous-time bound.
\end{restatable}

\paragraph{A posteriori discrete certificates.}
If a complete solver, LP relaxation, or dual method supplies any lower bound $B\le E_{\star}$, every decoded assignment $x$ has the certificate $0\le E(x)-E_{\star}\le E(x)-B$. Combining neural incumbents with classical lower bounds is therefore more informative than comparing final energies alone.

For explicit dense factors, one objective/gradient evaluation scales with the total table size $\sum_C q_C$, where $q_C=\prod_{i\in C}s_i$; the input alone requires $\Theta(\sum_C q_C)$ memory. Sparse, low-rank, Potts, cardinality, and logical factors can be cheaper when contracted by specialized kernels. Full derivations appear in Appendix~\ref{app:complexity}.

\section{Experiment}\label{sec:exp}

Throughout this section, \textsc{ReMAP} denotes the residual parameterization $z=u+\rho_v(G)$ of Eq.~\ref{eq:residual}. Energy and Loss report the best discrete incumbent and best relaxed objective over the run, respectively. rwise baselines are LBP, TRBP~\citep{1522634}, and Toulbar2~\citep{de2023toulbar2}; high-order comparisons use Toulbar2 and SRMP~\citep{6926846}. UAI files are unnormalized positive potentials converted by $\theta=-\log\phi$; the file layout is in Appendix~\ref{sec:read_uai}.

\subsection{Synthetic Problems} \label{sec:sys_exp}

Synthetic graphs are Erd\H{o}s--R\'enyi~\citep{Erdos:1959:pmd}. Pairwise and high-order instances use either Potts factors
\begin{equation}
    \theta_{ij} = \alpha\mathbb{I}(x_i = x_j) + \beta \label{eq:potts}
\end{equation}
or unary/pairwise values sampled uniformly from $[0.2,3.0]$. Nodes have $2$--$6$ labels; for Potts models, $\alpha,\beta\in[10^{-5},10^{3}]$. LBP/TRBP use $60$ iterations and damping $0.1$; Toulbar2 uses a $18000$s limit. Residual \textsc{ReMAP} uses a $5$-layer GraphSAGE branch with $d_l=1024$, learning rate $10^{-4}$, at most $150$ iterations, early stopping (absolute tolerance $10^{-4}$, patience $10$), and five random seeds. The two-layer theory in Section~\ref{sec:theory} is used only for the locality statement.

\textbf{Pairwise instances.} Table~\ref{table:pair_res_best} shows one small and one large graph for each energy model, using the best \textsc{ReMAP} energy over five seeds. The full best-of-five table is Table~\ref{table:pair_res_best_full} in Appendix~\ref{sec:full_syn_pair}; mean $\pm$ std over the same trials is Table~\ref{table:pair_res_full} in Appendix~\ref{sec:full_syn_pair}. On $1$k-node graphs the classical solvers are stronger or comparable; on the $50$k-node instances \textsc{ReMAP} attains the lowest energy on both models. The gap appears as size and density grow: residual lifting is most useful on the large, hard pairwise graphs where LBP, TRBP, and Toulbar2 degrade.

\begin{table}[t]
\centering
\caption{Representative pairwise synthetic instances. Discrete energy; lower is better. \textsc{ReMAP} is the best of five seeds. Full best-of-five and mean $\pm$ std tables are Table~\ref{table:pair_res_best_full} and Table~\ref{table:pair_res_full} in Appendix~\ref{sec:full_syn_pair}. Best in bold.}
\label{table:pair_res_best}
\resizebox{\columnwidth}{!}{
\begin{tabular}{cc|cccc}
\toprule
Graph & \#Nodes/\#Edges & LBP & TRBP & Toulbar2 & \textsc{ReMAP} \\
\midrule
P\_potts\_1 & 1k/7591 & $-22215.700$ & $-21365.800$ & $\mathbf{-22646.529}$ & $-21451.025$ \\
P\_potts\_5 & 50k/249624 & $25668.500$ & $35389.000$ & $12468.172$ & $\mathbf{11466.507}$ \\
P\_random\_1 & 1k/7540 & $\mathbf{-4901.100}$ & $-4505.020$ & $-4900.759$ & $-4564.763$ \\
P\_random\_5 & 50k/249374 & $12478.300$ & $13337.000$ & $12835.994$ & $\mathbf{11750.969}$ \\
\bottomrule
\end{tabular}
}
\end{table}

\textbf{Higher-order instances.} Table~\ref{table:high_res} reports three generated Potts high-order graphs. Clique counts include pairwise edges. \textsc{ReMAP} obtains lower energy than Toulbar2, which returns no solution within the time limit, and SRMP.

\begin{table}[t]
\centering
\caption{Synthetic high-order MRFs. Energy values; lower is better. ``NA'': no solution within the time limit. Best in bold.}
\label{table:high_res}
\resizebox{\columnwidth}{!}{
\begin{tabular}{cc|cccc}
\toprule
\multicolumn{1}{c}{\multirow{2}{*}{Graph}} & \multicolumn{1}{c|}{\multirow{2}{*}{\#Nodes/\#Cliques}} & \multicolumn{1}{c}{\multirow{2}{*}{Toulbar2}} & \multicolumn{1}{c}{\multirow{2}{*}{SRMP}} & \multicolumn{2}{c}{\textsc{ReMAP}} \\
\cline{5-6}
& & & & Energy & Loss \\
\midrule
H\_Instances\_1 & 500/41253 & NA & $-5785.093$ & $\mathbf{-7866.214\pm 389.207}$ & $-7859.68\pm 393.719$ \\
H\_Instances\_2 & 500/57934 & NA & $-18504.788$ & $\mathbf{-20260.289\pm 143.276}$ & $-20286.571\pm 143.624$ \\
H\_Instances\_3 & 1000/36993 & NA & $-5903.131$ & $\mathbf{-7232.648\pm 337.393}$ & $-7229.483\pm 336.218$ \\
\bottomrule
\end{tabular}
}
\end{table}

\subsection{UAI 2022 Inference Competition datasets}
We evaluate UAI 2022 MPE instances under the competition budgets of $1200$s and $3600$s. \textsc{ReMAP} uses an $8$-layer residual GNN, at most $100$ iterations, and embedding widths $\{64,512,1024,4096,8192\}$. Pairwise numbers are in Table~\ref{table:uai_res} in Appendix~\ref{sec:full_pair_uai}; the width sweep is Table~\ref{table:uai_full_res} in Appendix~\ref{sec:full_pair_uai}. On easy Segmentation instances Toulbar2 reports optima. On harder ProteinFolding and Grids instances \textsc{ReMAP} improves over LBP/TRBP but does not overtake Toulbar2. High-order numbers are in Table~\ref{table:uai_high_res} in Appendix~\ref{sec:full_pair_uai}: unlike Table~\ref{table:high_res}, Toulbar2 attains lower energy than \textsc{ReMAP} on all five reported UAI high-order instances at both budgets.

\subsection{Real-world Challenge: Physical Cell Identity}
PCI assignment in LTE/5G is converted to a pairwise MRF in Appendix~\ref{sec:trans_mip}. Table~\ref{table:pci_res} reports five real instances from a Chinese city, using the synthetic protocol with $100$ iterations and a $3600$s Toulbar2 limit. Toulbar2 is exact on the small graphs; \textsc{ReMAP} is strongest on PCI\_4 and PCI\_5. Five larger synthetic PCI instances are in Table~\ref{table:pci_synthetic} in Appendix~\ref{sec:trans_mip}.

\begin{table}[t]
\centering
\caption{Real PCI instances. Energy values; lower is better. Best in bold.}
\label{table:pci_res}
\resizebox{\columnwidth}{!}{
\begin{tabular}{cc|ccccc}
\toprule
\multicolumn{1}{c}{\multirow{2}{*}{Graph}} & \multicolumn{1}{c|}{\multirow{2}{*}{\#Nodes/\#Edges}} & \multicolumn{1}{c}{\multirow{2}{*}{LBP}} & \multicolumn{1}{c}{\multirow{2}{*}{TRBP}} & \multicolumn{1}{c}{\multirow{2}{*}{Toulbar2}} & \multicolumn{2}{c}{\textsc{ReMAP}} \\
\cline{6-7}
& & & & & Energy & Loss \\
\midrule
PCI\_1 & 30/165 & 20.344 & 20.455 & $\mathbf{18.134}$ & $18.372\pm 0.161$ & $18.373\pm 0.160$ \\
PCI\_2 & 40/311 & $\mathbf{98.364}$ & 98.762 & $\mathbf{98.364}$ & $98.555\pm 0.109$ & $98.555\pm 0.109$ \\
PCI\_3 & 80/1522 & $\mathbf{1003.640}$ & $\mathbf{1003.640}$ & $\mathbf{1003.640}$ & $\mathbf{1003.640\pm 0.0}$ & $1003.639\pm 0.0$ \\
PCI\_4 & 286/10714 & 585.977 & 585.977 & 426.806 & $\mathbf{410.945\pm 2.009}$ & $410.996\pm 2.014$ \\
PCI\_5 & 929/29009 & 1591.590 & 1591.590 & 1118.097 & $\mathbf{1074.617\pm 5.501}$ & $1074.676\pm 5.503$ \\
\bottomrule
\end{tabular}
}
\end{table}

\subsection{Analysis and Ablation Study}\label{sec:ablation}
Table~\ref{table:pci_ablation} compares residual $z=u+\rho_v(G)$, GNN-only $z=\rho_v(G)$, and free logits $z=u$ on the five real PCI instances, using $F(p)$, the same rounding, and---when a GNN is present---a $5$-layer GraphSAGE branch, $100$ iterations, and learning rate $10^{-4}$. Residual improves over GNN-only, which improves over free logits; the gap is concentrated on PCI\_4 and PCI\_5. This compares terminal discrete energies on three trajectories, not residual alignment $\beta_t$ in Theorem~\ref{thm:dyn}.

\begin{table}[t]
\centering
\caption{Parameterization ablation on the five real PCI instances. Discrete energy (mean $\pm$ std over five seeds); lower is better.}
\label{table:pci_ablation}
\resizebox{\columnwidth}{!}{
\begin{tabular}{cc|ccc}
\toprule
Instance & \#Nodes/\#Edges & Residual $u+\rho_v$ & GNN-only $\rho_v$ & Free logits $u$ \\
\midrule
PCI\_1 & 30/165    & $\mathbf{18.372 \pm 0.161}$ & $18.399 \pm 0.151$ & $18.773 \pm 0.000$ \\
PCI\_2 & 40/311    & $\mathbf{98.555 \pm 0.109}$ & $98.733 \pm 0.159$ & $98.874 \pm 0.000$ \\
PCI\_3 & 80/1522   & $\mathbf{1003.640 \pm 0.000}$ & $1004.789 \pm 1.241$ & $\mathbf{1003.640 \pm 0.000}$ \\
PCI\_4 & 286/10714 & $\mathbf{410.945 \pm 2.009}$ & $413.458 \pm 5.166$ & $448.274 \pm 0.000$ \\
PCI\_5 & 929/29009 & $\mathbf{1074.617 \pm 5.501}$ & $1092.796 \pm 13.663$ & $1286.412 \pm 0.000$ \\
\bottomrule
\end{tabular}
}
\end{table}

Anytime PCI comparison with Toulbar2 is Table~\ref{table:time} in Appendix~\ref{sec:fig_anaylsis}. Backbone and optimizer controls are Fig.~\ref{fig:GNN_backbone_curve} and Fig.~\ref{fig:optimizer} in Appendix~\ref{sec:fig_anaylsis}. Grid-graph wall-clock scaling is Fig.~\ref{fig:grid_time_vs_nodes} in Appendix~\ref{sec:fig_anaylsis}. Quality-versus-iteration curves are shown in Fig.~\ref{fig:qulity_eff} in Appendix~\ref{app:more_ana}; GNN-parameter landscapes are Fig.~\ref{fig:land} in Appendix~\ref{sec:fig_anaylsis}.

\section{Conclusion}\label{sec:conclusion}
Residual \textsc{ReMAP} composes an exact multilinear MAP objective with a graph-structured optimization geometry. The free-logit bypass gives a global normal form: neural lifting changes neither feasible logits nor the landscape, but joint optimization induces the adaptive metric $I+JJ^{\top}$. This yields exact dissipation and stationarity guarantees, four-hop locality for a two-layer branch, a saddle characterization of every finite critical point, and margin-dependent exact decoding and local convergence. Empirically, \textsc{ReMAP} is especially useful on larger, denser, and generated high-order instances, while Toulbar2 remains stronger on many small or standard UAI cases. On the five real PCI instances, a matched-protocol ablation of terminal discrete energy finds residual \textsc{ReMAP} improving over GNN-only and free logits, with the gap concentrated on the two larger cases. That comparison does not identify the squared alignment ratio $\beta_t=\|J_v^{\top}q\|_{2}^{2}/\|q\|_{2}^{2}$ of Theorem~\ref{thm:dyn}, which measures extra dissipation at a shared logit state rather than the endpoints of three different trajectories.

\section*{AI Use Disclosure}
In this work, we used generative AI tools for assisting with the verification of mathematical proofs. We have not used generative AI tools for any other tasks with required disclosure, and the remaining required disclosure tasks are not applicable to this work. Additionally, we used generative AI tools for editing the paper to improve readability and for editing software code. We have reviewed all AI-assisted work. In particular, the AI-assisted code modifications were verified and tested for correctness by 2 authors. We take responsibility for the final content of this work, including text, claims, or artifacts produced with the aid of generative AI.

\clearpage

\bibliography{iclr2027_conference}
\bibliographystyle{iclr2027_conference}

\newpage
\appendix
\onecolumn

\section{Related work}\label{app:related}

\textbf{Classical MAP inference.}
MAP inference in MRFs is NP-hard in general~\citep{koller2009probabilistic}. Practical methods include variants of belief propagation~\citep{article_lbp1,Felzenszwalb2004EfficientBP,article_lbp2,JMLR:v6:ihler05a}, tree-reweighted message passing~\citep{1522634,kolmogorov_convergent_2006,wainwright2005map}, SRMP for high-order MRFs~\citep{6926846}, and complete solvers such as Toulbar2~\citep{de2023toulbar2}. L\^{e}-Huu and Paragios~\citep{lehuu2018continuous} already established tightness of the same multilinear relaxation and proposed block-coordinate and ADMM algorithms. Exactness of $F(p)$ is therefore not a novelty of \textsc{ReMAP}; the contribution is the residual neural parameterization and its landscape/preconditioning analysis.

\textbf{Unsupervised GNNs for combinatorial optimization.}
Graph Neural Networks have demonstrated their power in optimization~\citep{yu2019learning,ying2024boosting}, with recent unsupervised GNN methods capturing combinatorial structure~\citep{GNN_co_1}. This line includes TSP~\citep{TSP1,TSP2}, VRP~\citep{wu2024neuralcombinatorialoptimizationalgorithms}, SAT~\citep{co_cite}, and Max Independent Set / Max Cut~\citep{Schuetz_2022}. PI-GNN~\citep{Schuetz_2022} optimizes a relaxed combinatorial objective through a GNN on each instance, primarily for binary QUBO and polynomial extensions. Residual \textsc{ReMAP} instead targets heterogeneous categorical variables with explicit arbitrary-order factors, and adds a free-logit residual so that the lifting is a global coordinate change rather than a GNN-only parameterization.

\textbf{Learning for MRFs.}
Supervised neural inference remains common. GNNs were used for MRF inference by~\citep{yoon2019inferenceprobabilisticgraphicalmodels} on graphs of up to $16$ nodes. Neural Enhanced BP~\citep{pmlr-v130-garcia-satorras21a}, BPNN~\citep{NEURIPS2020_07217414}, a neural max-product method~\citep{NEURIPS2020_61c66a2f}, and variational message-passing networks~\citep{pmlr-v180-cui22a} learn reusable inference operators or message updates across a distribution of instances. CRF-style combinations with GNNs have also been studied~\citep{pmlr-v97-qu19a,Tang_2021_CVPR}. These methods answer a different systems question from fitting a fresh optimizer to one instance.

Self-supervised alternatives include learning-to-optimize for MRFs~\citep{nair2021solvingmixedintegerprograms}, ALM-based CMPE models~\citep{pmlr-v238-arya24b}, SSMP~\citep{Arya_Rahman_Gogate_2024}, and SINE~\citep{arya2025sine}. SSMP combines a model/query predictor with inference-time refinement and teacher--student training; SINE improves structural embeddings and decoding for MPE. Consequently, neither ``a neural network optimizes the graphical-model objective'' nor ``no MAP labels are needed'' is a standalone novelty claim. Whether graph message passing, rather than the multilinear objective or node-specific parameters, improves optimization is tested by the residual / GNN-only / free-logit controls in Section~\ref{sec:ablation}.

Table~\ref{table:positioning} summarizes the methodological positioning.

\begin{table}[t]
\centering
\caption{Methodological positioning. ``Per instance'' means that optimization variables are fitted for the test instance.}
\label{table:positioning}
\resizebox{\textwidth}{!}{
\begin{tabular}{lcccc}
\toprule
Method & Use & Variables/factors & Search parameterization & Decoding \\
\midrule
Nonconvex MAP relaxation~\citep{lehuu2018continuous} & Per instance & Categorical; arbitrary order & Free product-simplex variables; multilinear & Guaranteed non-increasing block rounding \\
PI-GNN~\citep{Schuetz_2022} & Per instance & Primarily binary QUBO / polynomial & GNN optimizing a relaxed Hamiltonian & Projection to a discrete assignment \\
SSMP~\citep{Arya_Rahman_Gogate_2024} & Amortized + refinement & General PGM / MPE queries & Neural query predictor with self-supervised refinement & Threshold/argmax-style assignment \\
SINE~\citep{arya2025sine} & Amortized inference & MPE queries & Structure- and parameter-aware neural embeddings & Oracle-assisted or neighborhood decoding \\
Residual \textsc{ReMAP} (this work) & Per instance & Heterogeneous categorical; explicit arbitrary-order factors & Free logits plus a GNN residual on the multilinear energy & Conditional-expectation rounding \\
\bottomrule
\end{tabular}
}
\end{table}

\section{Padding procedure} \label{sec:padding}
\begin{figure}[tb]
  \centering
  \includegraphics[width=.75\linewidth]{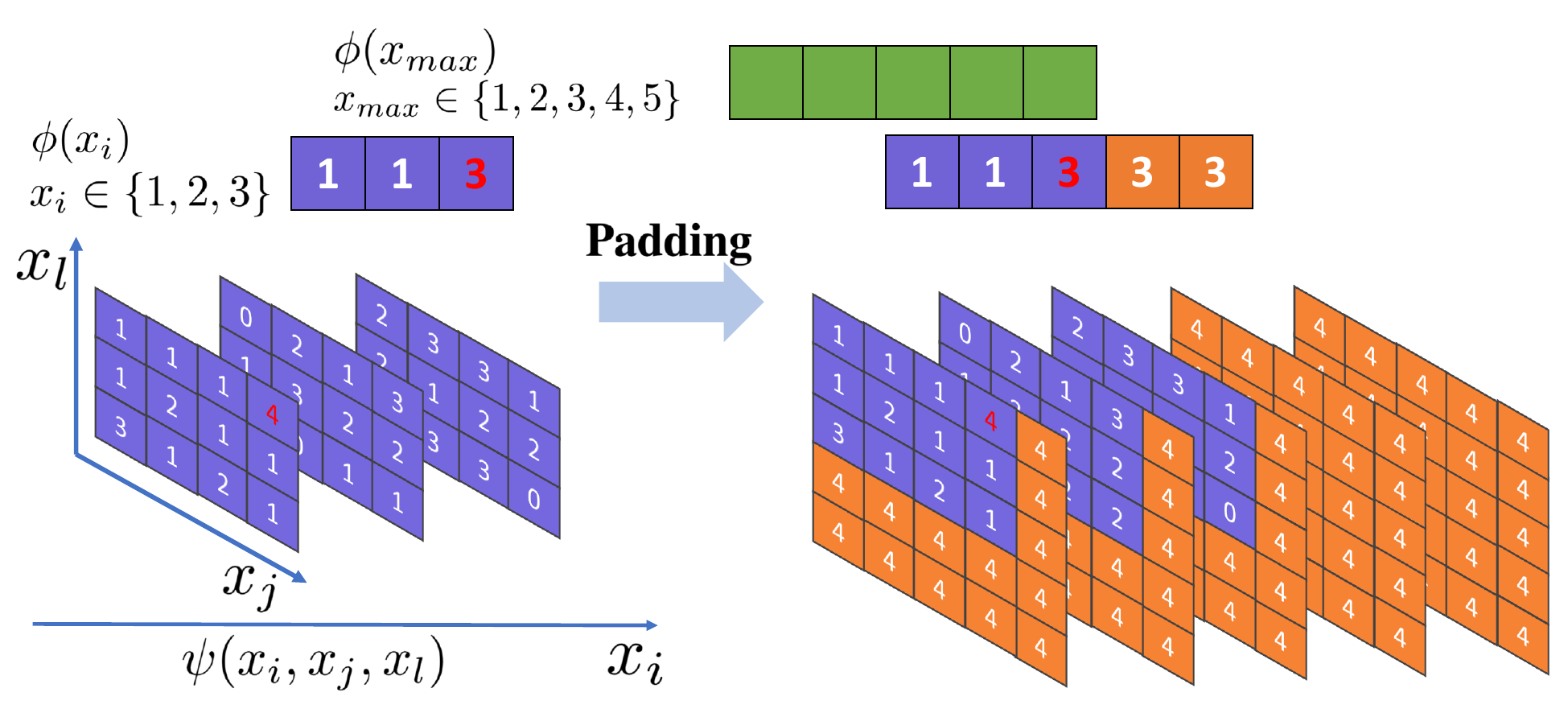}
  \caption{This illustrates the padding procedure for unary loss terms $\phi(x)$ and clique loss terms $\psi(x_i, x_j, x_k)$, with $|\mathcal{X}|=5$. $x_{max}$ denotes the variable that has the maximum value range.
  The elements shown in purple represent the energy values in the original $\phi$ and $\psi$. After padding, the dimension of vector $\phi$, as well as each dimension of the energy tensor $\psi(x_i, x_j, x_k)$, will be 5. Orange entries are a layout placeholder (implementation: $\max(\phi)$ or $\max(\psi)$); after $-\infty$ masking they do not enter $F(p)$.}
  \label{fig:padding}
\end{figure}

The schematic diagram of the padding procedure is in Fig.~\ref{fig:padding}. Padding equalizes tensor ranks so that a batched GNN can share one last dimension $|\mathcal{X}|$; it is not a substitute for masking. In this example, $|\mathcal{X}|=5$ and the unary energy of $x_i$ is $\phi(x_i)=\{1, 1, 3\}$. The implementation fills dummy slices with the original table maximum (here $3$), shown in orange. Clique tables are padded analogously: a $3\times3\times4$ factor becomes a $5\times5\times5$ tensor whose extra slices are filled with the original maximum (here $4$). After invalid logits are masked to $-\infty$, those placeholder entries are multiplied by zero probability and do not enter $F(p)$.

\section{Proofs of Section~\ref{sec:theory}} \label{app:proofs}
We use the Euclidean topology on the finite-dimensional polytope $\mathcal{P}$. For a label $a\in\mathcal{X}_i$, let $e_{i,a}\in\Delta_i$ denote the corresponding one-hot vector.

\subsection{Proof of Lemma~\ref{lem:block}}
\label{app:lemblock}
\begin{lemma}[Block-affine decomposition]
\label{lem:block}
For every $i\in\mathcal{V}$, every $p\in\mathcal{P}$, and every $a\in\mathcal{X}_i$,
$F(e_{i,a},p_{-i})=g_i(a;p_{-i})$ and $F(p)=\sum_{a}p_i(a)g_i(a;p_{-i})$.
In particular, $\min_a F(e_{i,a},p_{-i})\le F(p)$.
\end{lemma}
\begin{proof}
Fix $i$ and hold $p_{-i}$ fixed. Splitting the expectation according to the value of $X_i$ gives
\[
\mathbb{E}[E(X)]=\sum_{a}p_i(a)\,\mathbb{E}_{X_{-i}\sim\prod_{j\neq i}p_j}[E(a,X_{-i})]=\sum_{a}p_i(a)g_i(a;p_{-i}).
\]
Replacing $p_i$ by $e_{i,a}$ leaves $p_{-i}$ unchanged and forces $X_i=a$. The second identity is a convex combination of the vertex values.
\end{proof}

\subsection{Proof of Theorem~\ref{thm:identity}}
\label{app:identity}
\thmid*
\begin{proof}
Define the product mass $q_p(x)=\prod_i p_i(x_i)$. Expanding $E$ factor by factor and using independence yields $F(p)=\mathbb{E}[E(X)]$. Since $E(x)\ge E_{\star}$ for every $x$, averaging gives $F(p)\ge E_{\star}$. Conversely, the one-hot encoding of any MAP assignment $x^{\star}$ is a point mass, so $F(p)=E(x^{\star})=E_{\star}$. The gap decomposition $F(p)-E_{\star}=\sum_x q_p(x)(E(x)-E_{\star})$ has nonnegative summands, hence vanishes iff every $x$ in $\mathrm{supp}(q_p)$ is optimal. If $x^{\star}$ is unique, the support must be the singleton $\{x^{\star}\}$, so every marginal is the corresponding one-hot vector.
\end{proof}

\subsection{Proof of Corollary~\ref{cor:ce}}
\label{app:ce}
\corce*
\begin{proof}
Fix an arbitrary variable order $(i_1,\ldots,i_n)$. Set $p^{(0)}=p$. After $k-1$ blocks have been made one-hot, choose
$a_k\in\arg\min_a F(e_{i_k,a},p_{-i_k}^{(k-1)})$ and replace block $i_k$ by $e_{i_k,a_k}$. By Lemma~\ref{lem:block} the selected vertex value is no larger than the current convex combination, so $F(p^{(k)})\le F(p^{(k-1)})$. After $n$ steps, $p^{(n)}$ is the point mass at $x_{\mathrm{CE}}$ and $E(x_{\mathrm{CE}})=F(p^{(n)})\le F(p)$.
\end{proof}

\subsection{Proof of Proposition~\ref{prop:argmax}}
\label{app:argmax}
\propargmax*
\begin{proof}
If $Y$ is finite-valued and $h$ is real, $|h(y)-\mathbb{E}[h(Y)]|\le\mathrm{osc}(h)\Pr(Y\neq y)$. Apply this to $Y=X_C$, $y=\hat{x}_C$, and $h=\theta_C$, then sum over factors. Independence gives $\Pr(X_C=\hat{x}_C)=\prod_{i\in C}p_i(\hat{x}_i)$; the union bound yields the second inequality.
\end{proof}

\subsection{Proof of Theorem~\ref{thm:softmax}}
\label{app:softmax}
\thmsoftmax*
\begin{proof}
Finite softmax maps onto the relative interior of each simplex via $z_i(a)=\tau\log p_i(a)$, and cannot reach the boundary vertices. Interior product distributions can approach a MAP one-hot encoding, so $\inf_z F(p(z))=E_{\star}$ by continuity and Theorem~\ref{thm:identity}; the minimum of $F$ over the closed polytope $\mathcal{P}$ is attained at vertices, but that minimum is generally not attained at finite $z$. Every finite-logit product has full support, so $F(p)=E_{\star}$ is attained at finite logits iff every assignment is optimal. For the quantitative bound, the stated logits satisfy $1-p_i(x_i^{\star})\le(s_i-1)e^{-m/\tau}$; a union bound and $0\le E(x)-E_{\star}\le R_E$ give the claim.
\end{proof}

\subsection{Proof of Lemma~\ref{lem:logitgrad}}
\label{app:lemgrad}
\begin{lemma}[Logit-gradient identity]
\label{lem:logitgrad}
For finite $z$, node $i$, and label $b\in\mathcal{X}_i$,
$\partial\ell_{\tau}/\partial z_i(b)=(p_i(b)/\tau)\,(g_i(b;p_{-i})-F(p))$.
Consequently, $\nabla\ell_{\tau}(z)=0$ iff every $g_i(\cdot;p_{-i})$ is constant.
\end{lemma}
\begin{proof}
Use $\partial p_i(a)/\partial z_i(b)=\tau^{-1}p_i(a)(\mathbf{1}\{a=b\}-p_i(b))$ and $F=\sum_a p_i(a)g_i(a;p_{-i})$. Every finite-softmax probability is positive, so the block gradient vanishes exactly when all conditional costs equal their mean $F(p)$.
\end{proof}

\subsection{Proof of Theorem~\ref{thm:lossless}}
\label{app:lossless}
\thmlossless*
\begin{proof}
The stated inverse of $T$ is immediate, and
\[
DT=\begin{pmatrix}I&J_v\\0&I_m\end{pmatrix},\qquad
DT^{-1}=\begin{pmatrix}I&-J_v\\0&I_m\end{pmatrix}.
\]
Thus $T$ is a global diffeomorphism and substitution gives $L_{\tau}\circ T^{-1}(z,v)=\ell_{\tau}(z)$. In these coordinates the objective is independent of $v$. For fixed $v$, $u\mapsto u+\rho_v(G)$ is onto $\mathbb{R}^{D}$, proving equality of infima. Diffeomorphisms preserve neighborhoods, criticality, and local extrema, while $\{L_{\tau}\le c\}=T^{-1}(\{\ell_{\tau}\le c\}\times\mathcal{V})$. At a critical point, $\nabla\ell_{\tau}=0$, so second derivatives of the coordinate map disappear and $\nabla^2 L_{\tau}$ is congruent to $\mathrm{diag}(\nabla^2\ell_{\tau},0)$; Sylvester's law of inertia completes the proof.
\end{proof}

\subsection{Proof of Theorem~\ref{thm:dyn}}
\label{app:dyn}
\thmdyn*
\begin{proof}
The chain rule gives $\nabla_u L_{\tau}=q$ and $\nabla_v L_{\tau}=J_v^{\top}q$. Hence
\[
\|\nabla_{u,v}L_{\tau}\|_{2}^{2}=\|q\|_{2}^{2}+\|J_v^{\top}q\|_{2}^{2}=q^{\top}(I+J_v J_v^{\top})q
\]
and $\dot z=\dot u+J_v\dot v=-(I+J_v J_v^{\top})q$. Taking the inner product with $q$ proves Eq.~\ref{eq:dissipation}. Integration yields $\int_0^T\|q(t)\|_{2}^{2}\,dt\le L_{\tau}(0)-E_{\star}$; the minimum is no larger than the time average. Under Eq.~\ref{eq:align}, the integrand is at least $(1+\beta)\|q\|_{2}^{2}$, so the bound improves by $(1+\beta)^{-1}$.
\end{proof}

\subsection{Proof of Corollary~\ref{cor:disc}}
\label{app:disc}
\cordisc*
\begin{proof}
The descent lemma and $\eta H\le 1$ give $L_{\tau}(w_{k+1})\le L_{\tau}(w_k)-(\eta/2)\|\nabla L_{\tau}(w_k)\|_{2}^{2}$. By the chain rule, $\|\nabla L_{\tau}(w_k)\|_{2}^{2}=\|q_k\|_{2}^{2}+\|J_{v_k}^{\top}q_k\|_{2}^{2}$. Sum over $k$, use $L_{\tau}(w_K)\ge E_{\star}$, and bound the minimum by the average. Under Eq.~\ref{eq:align} the summand is at least $(1+\beta)\|q_k\|_{2}^{2}$, which improves the bound by $(1+\beta)^{-1}$.
\end{proof}

\subsection{Proof of Proposition~\ref{prop:fourhop}}
\label{app:fourhop}
\propfourhop*
\begin{proof}
The output at node $i$ depends only on embeddings in its two-hop neighborhood $B_2(i)$, so $\partial\rho_i/\partial e_{\ell}=0$ for $\ell\notin B_2(i)$. Therefore $(J_e J_e^{\top})_{ik}=\sum_{\ell}(\partial\rho_i/\partial e_{\ell})(\partial\rho_k/\partial e_{\ell})^{\top}$ vanishes when $B_2(i)\cap B_2(k)=\emptyset$, which holds if $\mathrm{dist}_G(i,k)>4$. Partitioning Jacobian columns by $v=(e,\omega)$ gives the metric decomposition.
\end{proof}

\subsection{Proof of Theorem~\ref{thm:saddle}}
\label{app:saddle}
\thmsaddle*
\begin{proof}
Modulo one additive gauge per node, finite softmax is a diffeomorphism onto $\mathrm{relint}(\mathcal{P})$, so it is equivalent to study $F$ at an interior $p$. By Lemma~\ref{lem:logitgrad}, criticality makes all first-order feasible derivatives vanish. For $U_i=\{h_i:\mathbf{1}^{\top}h_i=0\}$, multi-affinity gives the exact expansion
\[
F(p+h)-F(p)=\sum_{\emptyset\neq S\subseteq\mathcal{V}}A_S((h_i)_{i\in S}),
\]
with $A_S$ multilinear and all $|S|=1$ terms zero. If every remaining term vanished, the polynomial would be locally, hence globally, constant, making all vertex values $E(x)$ equal. Choose a nonzero term of smallest order $k\ge 2$ and tangent vectors of value $c\neq 0$. Perturbing only those $k$ blocks, lower-order terms vanish by minimality and higher-order terms require inactive blocks, so the change is exactly $ct^k$. Flipping one tangent vector changes the sign, and both higher and lower values occur arbitrarily near $p$. If the Hessian is nonzero, $k=2$ and the quadratic form takes both signs. Apply Theorem~\ref{thm:lossless} for the residual conclusions.
\end{proof}

\subsection{Proof of Theorem~\ref{thm:gap}}
\label{app:gap}
\thmgap*
\begin{proof}
The gap decomposition gives $\Delta(p)\ge\Gamma\Pr(X\neq x^{\star})$. If $\Delta<\Gamma$, conditional rounding has energy below $E_{\star}+\Gamma$ and must equal $x^{\star}$. If $\Delta<\Gamma/2$, then $q_p(x^{\star})>1/2$, so each marginal factor exceeds $1/2$ and $x_i^{\star}$ is the unique argmax.
\end{proof}

\subsection{Proof of Theorem~\ref{thm:flow}}
\label{app:flow}
\thmflow*
\begin{proof}
Let $\delta_i=1-p_i(x_i^{\star})$ and $S=\sum_i\delta_i$. Theorem~\ref{thm:gap} gives $\delta_i\le\Delta/\Gamma$, $S\le n\Delta/\Gamma$, and $\Delta\le R_E S$. For $a\neq x_i^{\star}$, let $A_i=\{X_{-i}\neq x_{-i}^{\star}\}$. Every complete-assignment energy lies in $[E_{\star},E_{\star}+R_E]$, so the range of $x_{-i}\mapsto E(a,x_{-i})$ is at most $R_E$ and
\[
g_i(a;p_{-i})-E(a,x_{-i}^{\star})\le R_E\Pr(A_i)\le R_E\sum_{j\neq i}\Pr(X_j\neq x_j^{\star})\le R_E S.
\]
The same bound holds for $a=x_i^{\star}$. Since $\kappa$ gives $E(a,x_{-i}^{\star})-E(x_i^{\star},x_{-i}^{\star})\ge\kappa$,
\[
g_i(a;p_{-i})-g_i(x_i^{\star};p_{-i})\ge\kappa-2R_E S.
\]
Inside the region $\Delta\le\varepsilon_0$, this is at least $\kappa/2$ and $p_i(x_i^{\star})\ge 1/2$. Lemma~\ref{lem:logitgrad} at the optimal label and $F-g_i(x_i^{\star};p_{-i})\ge(\kappa/2)\delta_i$ then yield
\[
\Bigl|\frac{\partial\ell_{\tau}}{\partial z_i(x_i^{\star})}\Bigr|=\frac{p_i(x_i^{\star})}{\tau}\bigl(F(p)-g_i(x_i^{\star};p_{-i})\bigr)\ge\frac{\kappa}{4\tau}\delta_i.
\]
Consequently
\[
\|q\|_{2}^{2}\ge\frac{\kappa^2}{16\tau^2}\sum_i\delta_i^2\ge\frac{\kappa^2 S^2}{16\tau^2 n}\ge\frac{\kappa^2}{16\tau^2 n R_E^2}\Delta^2=c_0\Delta^2.
\]
Dissipation keeps the trajectory in the region and gives $\dot\Delta\le -c_0\Delta^2$. Under Eq.~\ref{eq:align}, $\|\nabla_{u,v}L_{\tau}\|_{2}^{2}\ge(1+\beta)\|q\|_{2}^{2}$, so $c_0$ is replaced by $(1+\beta)c_0$. Integration proves the $O(1/t)$ rate. Exact conditional decoding follows from Theorem~\ref{thm:gap}; total variation from a point mass equals the mass outside that point.
\end{proof}

\subsection{Proof of Corollary~\ref{cor:gd}}
\label{app:gd}
\corgd*
\begin{proof}
The descent lemma, $\eta H\le 1$, and the chain rule give
\[
\Delta_{k+1}\le\Delta_k-\frac{\eta}{2}\bigl(\|q_k\|_{2}^{2}+\|J_{v_k}^{\top}q_k\|_{2}^{2}\bigr)\le\Delta_k-\frac{\eta c_0}{2}\Delta_k^2.
\]
Thus the objective cannot leave the small-gap sublevel set. If an iterate attains zero gap, the conclusion is immediate. Otherwise, taking reciprocals yields
\[
\frac{1}{\Delta_{k+1}}-\frac{1}{\Delta_k}=\frac{\Delta_k-\Delta_{k+1}}{\Delta_k\Delta_{k+1}}\ge\frac{\eta c_0}{2}.
\]
Telescoping proves the $O(1/k)$ rate. Under Eq.~\ref{eq:align} the decrement is multiplied by $1+\beta$, so $c_0$ is replaced by $(1+\beta)c_0$. Theorem~\ref{thm:gap} gives the decoding and total-variation conclusions.
\end{proof}

\subsection{Derivation of the a posteriori certificate}
\label{app:certificate}
Let $B$ be any valid lower bound, so $B\le E_{\star}\le E(x)$ for every decoded assignment $x$. Rearrangement gives $0\le E(x)-E_{\star}\le E(x)-B$. If $x=x_{\mathrm{CE}}$, Corollary~\ref{cor:ce} additionally gives $E(x_{\mathrm{CE}})\le F(p)$, which is an upper-bound comparison and does not replace an independently valid lower bound $B$.

\section{Relation to Lifting}\label{app:rel_lifting}
Our framework aligns with the classical concept of lifting~\citep{lift_cp_MIP} by expanding the parameter space while leaving the logit space unchanged. The residual map $T(u,v)=(u+\rho_v(G),v)$ is a global diffeomorphism, so the lifted objective is exactly the free-logit objective times a parameter fiber (Theorem~\ref{thm:lossless}). The GNN does not create a larger set of logits or a low-energy tunnel in the discrete space; it changes the optimization metric to $I+JJ^{\top}$.

\section{Complexity Analysis}\label{app:complexity}
For a dense explicit factor $C$, evaluating its contribution to Eq.~\ref{eq:multilinear} enumerates $q_C=\prod_{i\in C}s_i$ table entries. Summing over factors gives the forward arithmetic bound $T_{\mathrm{factor}}=O(\sum_C q_C)$. Reverse-mode differentiation has the same asymptotic arithmetic order. Since the dense tables themselves contain exactly $\sum_C q_C$ scalars, their memory lower bound is $\Theta(\sum_C q_C)$. If $s_i\le s_{\max}$ and $|C|\le r$, then $q_C\le s_{\max}^{r}$ and the cost is $O(|\mathcal{C}|s_{\max}^{r})$; it is not $O(r|\mathcal{C}|s_{\max})$.

Representation structure can reduce this cost. A sparse table with a default value plus $m_C$ exceptional tuples is $O(m_C|C|)$ after the default expectation. A CP factorization $\theta_C(x_C)=\sum_{h=1}^{R_C}\prod_{i\in C}a_{Cih}(x_i)$ contracts in $O(R_C\sum_{i\in C}s_i)$ plus product operations. Potts, cardinality, and logical factors likewise admit specialized kernels.

Let $d_h$ be hidden width, $K$ the number of layers, and $E_G$ the clique-expanded edge set. One dense GraphSAGE layer costs approximately $O(|\mathcal{V}|d_h^{2}+|E_G|d_h)$, and storing activations across $K$ layers costs $O(K|\mathcal{V}|d_h)$ before checkpointing. Conditional-expectation rounding need not recompute the full energy for every candidate label: caching incident-factor contractions gives a worst-case sweep cost $O(\sum_C |C|q_C)$ for dense factors.

\section{Full results on Pairwise Synthetic instances} \label{sec:full_syn_pair}
This appendix expands the pairwise synthetic comparison of Section~\ref{sec:sys_exp}. Instances are Erd\H{o}s--R\'enyi graphs with prefix ``P\_potts\_'' for Potts pairwise energies (Eq.~\ref{eq:potts}) and ``P\_random\_'' for unary and pairwise values sampled uniformly from $[0.2,3.0]$. Node cardinalities range from $2$ to $6$. LBP and TRBP run for $60$ iterations with damping $0.1$; Toulbar2 uses a default mode with a $18000$s limit. Residual \textsc{ReMAP} uses a $5$-layer GraphSAGE branch, $d_l=1024$, learning rate $10^{-4}$, at most $150$ iterations, and early stopping (absolute tolerance $10^{-4}$, patience $10$).

Table~\ref{table:pair_res_best} in the main text reports four representative graphs. Table~\ref{table:pair_res_best_full} lists the best discrete energy among the same five \textsc{ReMAP} seeds on all eighteen instances. Table~\ref{table:pair_res_full} reports mean $\pm$ standard deviation of Energy and Loss over those five trials. Energy is the decoded discrete energy; Loss is the relaxed multilinear objective. Lower is better for both.

On $1$k--$10$k-node graphs, LBP or Toulbar2 typically attains the lowest energy, and the best \textsc{ReMAP} seed remains comparable on Potts instances. From $50$k nodes onward, the best \textsc{ReMAP} seed is lowest on most Potts and random instances in Table~\ref{table:pair_res_best_full}. The mean in Table~\ref{table:pair_res_full} is more conservative: several dense $50$k-node graphs have a large seed standard deviation, so a single best seed in Table~\ref{table:pair_res_best_full} should not be read as a tight typical value.

\begin{table*}[tb]
\begin{center}
\caption{Full pairwise synthetic results. Discrete energy; lower is better. \textsc{ReMAP} is the best energy over five seeds. The four-row excerpt in the main text is Table~\ref{table:pair_res_best}. Best in bold.}
\label{table:pair_res_best_full}
\scalebox{0.49}{
\begin{tabular}{cc|cccc||cc|cccc}
\toprule
Graph & \multicolumn{1}{c|}{\#Nodes/\#Edges} & LBP  & TRBP & Toulbar2  & \textsc{ReMAP}  & Graph & \multicolumn{1}{c|}{\#Nodes/\#Edges} & LBP  & TRBP & Toulbar2  & \textsc{ReMAP}  \\
\midrule
P\_potts\_1 & 1k/7591  & -22215.700   & -21365.800    & \textbf{-22646.529}      & -21451.025 & P\_random\_1 & 1k/7540  & \textbf{-4901.100}    & -4505.020     & -4900.759     &  -4564.763 \\
P\_potts\_2 & 5k/37439  & \textbf{-111319.000}   & -105848.000    & -110022.248      & -105952.531 & P\_random\_2 & 5k/37488  & -24059.900     & -22934.000      & \textbf{-24139.194}     &  -21834.693 \\
P\_potts\_3 & 10k/75098  & \textbf{-221567.000}   & -210570.000    & -218311.424      & -209925.269 & P\_random\_3 & 10k/74518  & -47873.200    & -47002.000     & \textbf{-48107.172}     &  -42120.325 \\
P\_potts\_4 & 50k/248695  & 12411.200   & 13454.600    & 12955.129      & \textbf{11679.429} & P\_random\_4 & 50k/249554  & 12881.500    & 14342.300     & 12233.890     &  \textbf{11769.934} \\
P\_potts\_5 & 50k/249624  & 25668.500   & 35389.000    & 12468.172      &  \textbf{11466.507}  & P\_random\_5 & 50k/249374  & 12478.300    & 13337.000    & 12835.994      &  \textbf{11750.969} \\
P\_potts\_6 & 50k/300181  &  17609.800        &  17362.600         & 17635.791      & \textbf{16756.999}  & P\_random\_6 & 50k/299601  & 16723.600    &  16754.500         & 18031.964      & \textbf{16700.674}  \\
P\_potts\_7 & 50k/299735  &   \textbf{16962.500}       &  \textbf{16962.500}     & 19532.817        &  17002.578   &  P\_random\_7 & 50k/299538  &  \textbf{16689.200}   & 16701.600        & 18179.548       &  \textbf{16689.252}    \\
P\_potts\_8 & 50k/374169  &   \textbf{24552.400}      &   24596.800         & 25446.235       &   \textbf{24552.413}  & P\_random\_8 & 50k/374203  & \textbf{24556.000}       & \textbf{24556.000}          & 25549.594      &   \textbf{24555.995} \\
P\_potts\_9 & 50k/375603  & 25099.800      &  25095.600          & 25502.495      &  \textbf{25050.522}  & P\_random\_9 & 50k/374959 & \textbf{24635.600}      & 24689.500          & 25908.500     & \textbf{24640.039}    \\
\bottomrule
\end{tabular}
}
\end{center}
\end{table*}

\begin{table*}[tb]
\begin{center}
\caption{Pairwise synthetic instances, same five trials as Table~\ref{table:pair_res_best_full}. \textsc{ReMAP} Energy and Loss are mean $\pm$ std. Energy is the decoded discrete energy; Loss is the relaxed multilinear objective. Lower is better.}
\label{table:pair_res_full}
\scalebox{0.7}{
\begin{tabular}{cc|ccccc}
\toprule
\multicolumn{1}{c}{\multirow{2}{*}{Graph}} & \multicolumn{1}{c|}{\multirow{2}{*}{\#Nodes/\#Edges}} & \multicolumn{1}{c}{\multirow{2}{*}{LBP}}    &\multicolumn{1}{c}{\multirow{2}{*}{TRBP}}  &\multicolumn{1}{c}{\multirow{2}{*}{Toulbar2}} &  \multicolumn{2}{c}{\textsc{ReMAP}}           \\  \cline{6-7}
\multicolumn{1}{c}{}                       & \multicolumn{1}{c|}{}       & \multicolumn{1}{c}{}    & \multicolumn{1}{c}{}  & \multicolumn{1}{c}{}  & \multicolumn{1}{c}{Energy} & Loss \\ 
\midrule   
P\_potts\_1 & 1k/7591  & -22215.700   & -21365.800    & -22646.529      & -21791.868 $\pm$ 218.106  & -21799.268 $\pm$ 216.075 \\
P\_potts\_2 & 5k/37439  & -111319.000   & -105848.000    & -110022.248      & -105762.092 $\pm$ 434.674    & -106016.855 $\pm$ 168.861 \\
P\_potts\_3 & 10k/75098  & -221567.000   & -210570.000    & -218311.424      & -211406.914  $\pm$ 1489.099    &  -210182.681 $\pm$ 275.164 \\
P\_potts\_4 & 50k/248695  & 12411.200   & 13454.600    & 12955.129      & 12219.817  $\pm$ 538.969  & 11811.682  $\pm$ 123.877 \\
P\_potts\_5 & 50k/249624  & 25668.500   & 35389.000    & 12468.172      &  12010.036 $\pm$ 266.610   & 11673.708 $\pm$ 146.513  \\
P\_potts\_6 & 50k/300181  &  17609.800        &  17362.600         & 17635.791      & 18399.913 $\pm$ 1475.526    & 16988.347 $\pm$ 163.587  \\
P\_potts\_7 & 50k/299735  &   16962.500       &  16962.500     & 19532.817        &  17480.701 $\pm$ 212.084    & 17265.434 $\pm$ 140.904    \\
P\_potts\_8 & 50k/374169  &   24552.400      &   24596.800         & 25446.235       &   26115.840 $\pm$ 1677.627   &  24668.087 $\pm$ 163.587  \\
P\_potts\_9 & 50k/375603  & 25099.800      &  25095.600          & 25502.495      &  26348.525 $\pm$ 1095.319   & 25189.789 $\pm$ 132.693  \\ 

\midrule 
P\_random\_1 & 1k/7540  & -4901.100    & -4505.020     & -4900.759     &  -4570.079 $\pm$ 31.228 & -4574.664 $\pm$ 31.411\\
P\_random\_2 & 5k/37488  & -24059.900     & -22934.000      & -24139.194     &  -21774.416 $\pm$ 52.910 & -21798.702 $\pm$ 32.389 \\
P\_random\_3 & 10k/74518  & -47873.200    & -47002.000     & -48107.172     &  -41953.991  $\pm$ 237.577 & -41972.379 $\pm$ 216.574 \\
P\_random\_4 & 50k/249554  & 12881.500    & 14342.300     & 12233.890     &  12552.252 $\pm$ 30.311 & 11983.388$\pm$ 213.454 \\
P\_random\_5 & 50k/249374  & 12478.300    & 13337.000    & 12835.994      &  12308.580 $\pm$ 14.045 & 11945.450 $\pm$ 194.481 \\
P\_random\_6 & 50k/299601  & 16723.600    &  16754.500         & 18031.964      & 17705.219 $\pm$ 435.560  & 17207.997 $\pm$ 405.217  \\
P\_random\_7 & 50k/299538  &  16689.200   & 16701.600        & 18179.548       &  18343.026 $\pm$ 1448.821 & 16971.435 $\pm$ 209.021    \\
P\_random\_8 & 50k/374203  & \textbf{24556.000}       & \textbf{24556.000}          & 25549.594      &   25949.446 $\pm$ 995.956 & 24787.343 $\pm$ 163.587 \\ 
P\_random\_9 & 50k/374959 & \textbf{24635.600}      & 24689.500          & 25908.500     & 25871.264 $\pm$ 1087.915 &  24811.354 $\pm$ 171.315    \\
\bottomrule
\end{tabular}   
}
\end{center}

\end{table*}

\section{Results of UAI Inference Competition 2022 dataset} \label{sec:full_pair_uai}
This appendix reports the UAI 2022 MPE numbers summarized in Section~\ref{sec:exp}. Pairwise and high-order instances use the competition budgets of $1200$s and $3600$s. LBP and TRBP run for $30$ iterations with damping $0.1$; Toulbar2 is limited to $1200$s in the pairwise table. Residual \textsc{ReMAP} uses an $8$-layer GraphSAGE branch, at most $100$ iterations, and the same UAI-file conversion as Appendix~\ref{sec:read_uai}. Unless noted, the embedding width is $1024$; Table~\ref{table:uai_full_res} varies that width.

Table~\ref{table:uai_res} lists pairwise instances. On Segmentation graphs Toulbar2 reports certified optima (marked ``opt''), and LBP/TRBP are already close; \textsc{ReMAP} is comparable but not better. On ProteinFolding and Grids instances, which are larger or denser, \textsc{ReMAP} improves over LBP and TRBP while remaining above Toulbar2. This is the intended reading of the main-text claim: residual lifting is more useful when message passing degrades, not a replacement for a complete solver on small, well-solved UAI cases.

Table~\ref{table:uai_high_res} lists five large high-order competition instances. Unlike the generated high-order graphs in Table~\ref{table:high_res}, Toulbar2 attains lower energy than \textsc{ReMAP} at both $1200$s and $3600$s on every reported instance. The shift should be read together with factor order, table representation, and energy range rather than as a uniform high-order advantage.

Table~\ref{table:uai_full_res} repeats the pairwise suite across embedding widths $\{64,512,1024,4096,8192\}$. Width changes the attained energy, but a moderate width is already competitive on most graphs; larger embeddings are not uniformly better and cost more memory. The main experiments therefore keep a single width per suite rather than selecting the best column after the fact. 

\begin{table*}[ht]
\begin{center}
\caption{Results on the UAI inference competition 2022. Numbers correspond to the obtained energy values. Best in bold.``opt" denotes it is the optimal solution.}
\label{table:uai_res}
\scalebox{0.8}{
\setlength{\tabcolsep}{1.5pt}
\begin{tabular}{cc|ccccc}
\toprule

\multicolumn{1}{c}{\multirow{2}{*}{Graph}} & \multicolumn{1}{c|}{\multirow{2}{*}{\#Nodes/\#Edges}} & \multicolumn{1}{c}{\multirow{2}{*}{LBP}}    &\multicolumn{1}{c}{\multirow{2}{*}{TRBP}}  &\multicolumn{1}{c}{\multirow{2}{*}{Toulbar2}} &  \multicolumn{2}{c}{\textsc{ReMAP}}           \\  \cline{6-7}
\multicolumn{1}{c}{}                       & \multicolumn{1}{c|}{}       & \multicolumn{1}{c}{}    & \multicolumn{1}{c}{}  & \multicolumn{1}{c}{}  & \multicolumn{1}{c}{Energy} & Loss \\ 
\midrule
ProteinFolding\_11    &          400/7160        & -3106.080      & -3079.030          & -4461.047       & \multicolumn{1}{c}{-3976.908 $\pm$ 52.047}  &  -4018.784 $\pm$ 36.491       \\
ProteinFolding\_12     &          250/1848       &  3570.210    &   3604.240         & 3562.387(opt)           & \multicolumn{1}{c}{16137.682 $\pm$ 16.020}  & 16090.801 $\pm$ 22.869              \\
Grids\_19     &          1600/3200               & -2250.440     & -2103.610          & -2643.107          & \multicolumn{1}{c}{-2400.251 $\pm$ 20.061} & -2398.078 $\pm$ 16.010           \\
Grids\_21     &          1600/3200               &  -13119.300    & -12523.300            & -18895.393            &  \multicolumn{1}{c}{-16592.926 $\pm$ 94.368} &  -16605.564 $\pm$ 113.096          \\
Grids\_24     &          1600/3120               &  -13210.400    & -13260.900            & -18274.302             & \multicolumn{1}{c}{-16323.767 $\pm$ 171.950} & -16222.104 $\pm$ 222.593             \\
Grids\_25     &          1600/3120               & -2170.890     & -2171.050           & -2620.268            & \multicolumn{1}{c}{-2361.900 $\pm$ 10.231}  & -2361.055 $\pm$ 12.678            \\
Grids\_26     &          400/800                 &  -2063.350    &  -1903.910           & -3010.719             & \multicolumn{1}{c}{-2595.041 $\pm$ 43.306} &   -2577.378 $\pm$ 39.370        \\
Grids\_27     &          1600/3120               & -9024.640     & -9019.470            & -12284.284             &\multicolumn{1}{c}{-10898.595 $\pm$ 160.435} & -10771.257 $\pm$ 170.329  \\
Grids\_30     &          400/760                 & -2142.890     & -2154.910            & -2984.248             & \multicolumn{1}{c}{-2651.035 $\pm$ 35.508} & -2676.246 $\pm$ 19.886              \\
Segmentation\_11     &           228/624         & 329.950    &  339.762         & 312.760 (opt)             & \multicolumn{1}{c}{432.291 $\pm$ 34.208} & 391.971 $\pm$ 40.099             \\
Segmentation\_12     &           231/625         & 75.867     & 77.898         & 51.151 (opt)             & \multicolumn{1}{c}{90.248 $\pm$ 22.655} & 105.639 $\pm$ 21.165             \\
Segmentation\_13     &           225/607         & 75.299     &  88.554            & 49.859 (opt)             & \multicolumn{1}{c}{80.156 $\pm$ 6.462} & 78.685 $\pm$ 18.546          \\
Segmentation\_14     &           231/632         & 95.619    & 98.691           & 92.334 (opt)            & \multicolumn{1}{c}{102.263 $\pm$ 7.169} & 101.268 $\pm$ 5.467            \\
Segmentation\_15     &           229/622         & 412.990     & 418.853            & 380.393 (opt)             & \multicolumn{1}{c}{417.276 $\pm$ 22.357} & 408.214 $\pm$ 27.037              \\
Segmentation\_16     &           228/610         & 100.853    & 101.670            & 95.000 (opt)             & \multicolumn{1}{c}{102.687 $\pm$ 4.571} & 101.687 $\pm$ 7.358             \\
Segmentation\_17     &           225/612         & 421.888     & 432.012           & 407.065 (opt)             & \multicolumn{1}{c}{445.843 $\pm$ 24.459} & 478.881 $\pm$ 32.824            \\
Segmentation\_18     &           235/647         & 100.389     & 98.411            & 82.669 (opt)             & \multicolumn{1}{c}{104.721 $\pm$ 6.489} & 96.315 $\pm$ 6.124            \\
Segmentation\_19     &           228/624         & 86.589    &  86.692          &  58.704 (opt)            &    \multicolumn{1}{c}{96.173 $\pm$ 4.731} & 84.882 $\pm$ 10.063           \\
Segmentation\_20     &        232/635            & 289.435     &   291.527          & 262.216 (opt)             & \multicolumn{1}{c}{335.245 $\pm$ 36.163} & 315.482 $\pm$ 24.268    \\
\bottomrule
\end{tabular}
}
\end{center}
\end{table*}

\begin{table*}[ht] 
\begin{center}
\caption{Results on high-order cases of the UAI inference competition 2022. Numbers correspond to the obtained energy values. Best in bold.}
\label{table:uai_high_res}
\resizebox{\textwidth}{!}{
\begin{tabular}{cc|cccc}
\toprule
\multicolumn{1}{c}{\multirow{2}{*}{Graph}} & \multicolumn{1}{c|}{\multirow{2}{*}{\#Nodes/\#Cliques}} & \multicolumn{1}{c}{\multirow{2}{*}{Toulbar2 (1200s)}}    &\multicolumn{1}{c}{\multirow{2}{*}{Toulbar2 (3600s)}}   &  \multicolumn{2}{c}{\textsc{ReMAP}}           \\  \cline{5-6}
\multicolumn{1}{c}{}                       & \multicolumn{1}{c|}{}       & \multicolumn{1}{c}{}    & \multicolumn{1}{c}{}   & \multicolumn{1}{c}{Energy} & Loss \\ 
\midrule
Maxsat\_gss-25-s100   &   31931/96111                                   & \textbf{-145969.060}     &  -145969.060       & \multicolumn{1}{c}{-139904.266 $\pm$ 1717.483} & -139914.341 $\pm$ 1710.139 \\
BN-nd-250-5-10       &   250/250                                        &  155.129        &  \textbf{154.610}           & \multicolumn{1}{c}{189.395 $\pm$ 5.424} & 187.729 $\pm$ 4.966 \\
Maxsat\_mod4block\_2vars\_10gates\_u2\_autoenc & 479/123509             &  -186103.111    & -186103.111        & \multicolumn{1}{c}{-146166.620 $\pm$ 41250.035} & -146166.797 $\pm$ 41249.859\\
Maxsat\_mod2c-rand3bip-sat-240-3.shuffled-as.sat05-2520  & 339/2416     & -3734.627       &  \textbf{-3737.076}         & \multicolumn{1}{c}{-3511.994 $\pm$ 220.654} & -3511.822 $\pm$ 220.471  \\
Maxsat\_mod2c-rand3bip-sat-250-3.shuffled-as.sat05-2535  & 352/2492     & \textbf{-3863.259}       &  -3863.259         & \multicolumn{1}{c}{-3686.567 $\pm$ 166.998} & -3686.085 $\pm$ 166.499  \\


\bottomrule
\end{tabular}
}
\end{center}
\end{table*}

\begin{table}[ht]
\begin{center}
\caption{Full results on the UAI inference competition 2022 of \textsc{ReMAP} with different feature dimensions. Numbers correspond to the obtained energy values. }
\label{table:uai_full_res}
\resizebox{\textwidth}{!}{
\begin{tabular}{cc|ccccc}
\toprule
Graph & \multicolumn{1}{c|}{\#Nodes/\#Edges} & dim=64 & dim=512 & dim=1024   & dim=4096 & dim=8192\\ 
\midrule
ProteinFolding\_11    &          400/7160      & -3892.949  & -3886.701 & -3946.168      & -4065.294    & -4003.323      \\
ProteinFolding\_12     &          250/1848     &16064.795  &16068.406  &16051.798  & 16088.073             & 16071.324      \\
Grids\_19     &          1600/3200             &-2355.159  &-2404.975    &-2337.281     &  -2341.2746      & -2373.618   \\
Grids\_21     &          1600/3200             &-16478.466    &-16169.0320   & -16446.410 & -16209.017   &  -16278.668          \\
Grids\_24     &          1600/3120             &-16008.008    & -15900.249   & -15841.799   &-15608.162   & -15948.219       \\
Grids\_25     &          1600/3120             &-2343.547     &-2353.223  & -2319.899     & -2306.686   & -2288.182     \\
Grids\_26     &          400/800               &-2532.837     &-2608.395   & -2553.781     & -2559.572   & -2535.464          \\
Grids\_27     &          1600/3120             &-10748.024   & -10704.057     &-10514.857    & -10389.031    & -10665.737            \\
Grids\_30     &          400/760               &-2563.274     & -2631.862     & -2640.044  & -2691.091  &-2649.462            \\
Segmentation\_11     &           228/624         &330.541     &349.906        & 334.882    & 356.895             & 337.312            \\
Segmentation\_12     &           231/625         & 74.705     & 74.029        & 155.062    & 79.151             & 105.801          \\
Segmentation\_13     &           225/607         &67.371      & 86.064        & 69.430     & 72.394             & 112.516           \\
Segmentation\_14     &           231/632         & 94.192     & 96.501        & 100.582    & 104.091            &  96.572            \\
Segmentation\_15     &           229/622         & 388.223    & 386.701       & 397.246    & 407.731            & 390.641          \\
Segmentation\_16     &           228/610         & 99.086     & 99.690        & 111.121    & 98.209             & 108.360          \\
Segmentation\_17     &           225/612         &424.686     & 426.130       & 425.192    & 425.240            & 427.810            \\
Segmentation\_18     &           235/647         & 89.905     & 101.307       & 94.224     & 88.854             & 88.809          \\
Segmentation\_19     &           228/624         & 76.244     & 78.337        & 74.284     & 69.116             & 70.770    \\
Segmentation\_20     &        232/635            & 298.802    &  301.802      & 302.673    &  304.457           & 312.970 \\
\bottomrule
\end{tabular}
}
\end{center}
\end{table}

\section{More analysis}\label{app:more_ana}
This section supplements the ablation paragraph in Section~\ref{sec:ablation}. The parameterization table itself remains in the main text (Table~\ref{table:pci_ablation}); here we only discuss iteration-wise quality and, in Appendix~\ref{sec:fig_anaylsis}, landscapes, backbones, optimizers, anytime PCI traces, and grid scaling.
\begin{figure}[ht]
    \centering
    \includegraphics[width=1.0\linewidth]{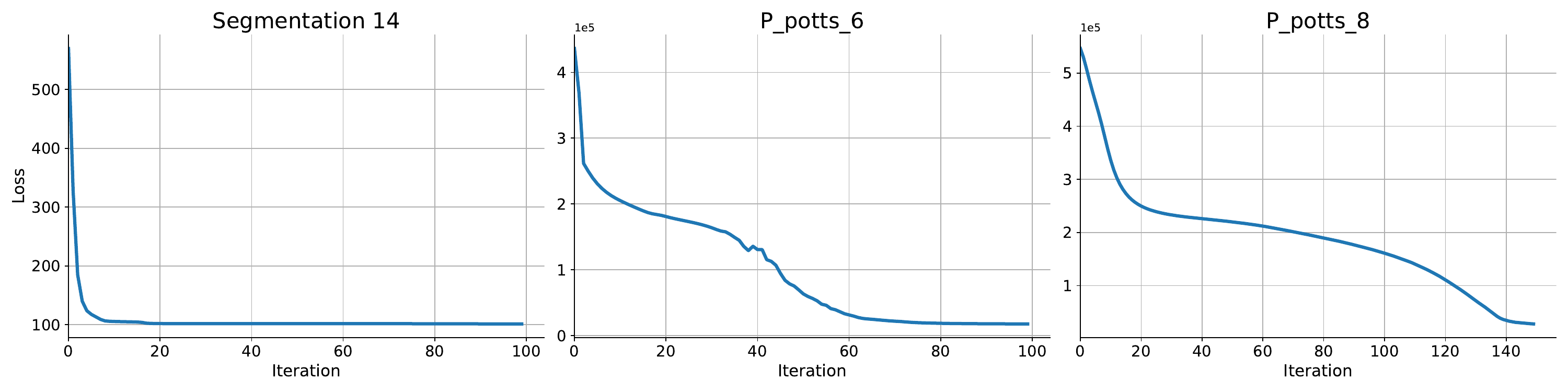}
    \caption{The loss curves of the Segmentation\_14, P\_potts\_6  and P\_potts\_8 from pairwise potts synthetic problems.}
    \label{fig:qulity_eff}
\end{figure}
\textbf{Efficiency vs Solution Quality.}
We evaluate the performance of the \textsc{ReMAP} using the same network size and a consistent learning rate of 1e{-4} on the Segmentation\_14 dataset from the UAI 2022 inference competition, along with two of our generated Potts instances: P\_potts\_6 and P\_potts\_8. This setup allows us to observe the trends associated with changes in graph size and sparsity. The results are presented in Fig.~\ref{fig:qulity_eff}. It is seen that the model converges rapidly when the graph is small and sparse, within approximately 20 iterations on the Segmentation\_14 dataset. Comparing P\_potts\_6 and P\_potts\_8, we observe that though both graphs are of the same size, the denser graph raises significantly more challenges during optimization. This indicates that increased size and density can complicate the optimization process, and \textsc{ReMAP} would need more time to navigate a high quality solution under such cases.

\section{GNN formulations} \label{sec:GNNs}
We summarized the popular GNN message passing formats in Table~\ref{table:GNNs} to show the logic behind the GNN backbone selection of our work.

\begin{table*}
    \caption{Graph convolutions in typical GNNs}
    \centering
    \begin{tabular}{ccc}
    \toprule
      & Graph Convolutions & Neighbor Influence  \\
      \hline
      GCN   & $h_i^{(k)}=\sigma \left(W_k \cdot \sum_{j \in \mathcal{N}(i) \cup \{i\}}(\operatorname{deg}(i) \operatorname{deg}(j))^{-1 / 2} h_j^{(k-1)}\right)$ & \textbf{Unequal}\\
        GAT  & $h_i^{(k)}=\sigma \left(\sum_{j \in \mathcal{N}(i) \cup\{i\}} \alpha_{i, j} W_k h_j^{(k-1)} \right)$ & \textbf{Unequal}\\
        GraphSAGE  & $h_i^{(k)}=\sigma\left(A_k h_i^{(k-1)} + B_k |\mathcal{N}(i)|^{-1}\sum_{j\in \mathcal{N}(i)} h_j^{(k-1)} + b_k\right)$ & \textbf{Equal}\\
    \bottomrule
    \end{tabular}
  \label{table:GNNs}
\end{table*}


\textbf{Loss Landscape Visualization.}
We visualize loss landscapes using the tool from \citep{visualloss}, with detailed settings in Appendix~\ref{app:vis_setup}. Fig.~\ref{fig:land} in Appendix~\ref{sec:fig_anaylsis} shows slices for networks of depths $K \in \{1,2,5,8\}$, while Fig.~\ref{fig:loss_layer} displays converged loss trends. These plots are in the GNN-parameter coordinates, not in logit space. Theorem~\ref{thm:lossless} already implies that residual lifting does not enlarge the feasible logit set or create new discrete solutions; depth only changes how the graph branch parameterizes $\rho_v(G)$ and therefore the induced metric $I+JJ^{\top}$. The two-layer four-hop locality result does not apply to the deeper experimental stacks, which are retained as the original training architecture.

\section{Results about the analysis experiments}\label{sec:fig_anaylsis}

\begin{figure}[ht]
    \centering
    \begin{minipage}[t]{0.88\textwidth} 
        \centering
        \includegraphics[width=\linewidth]{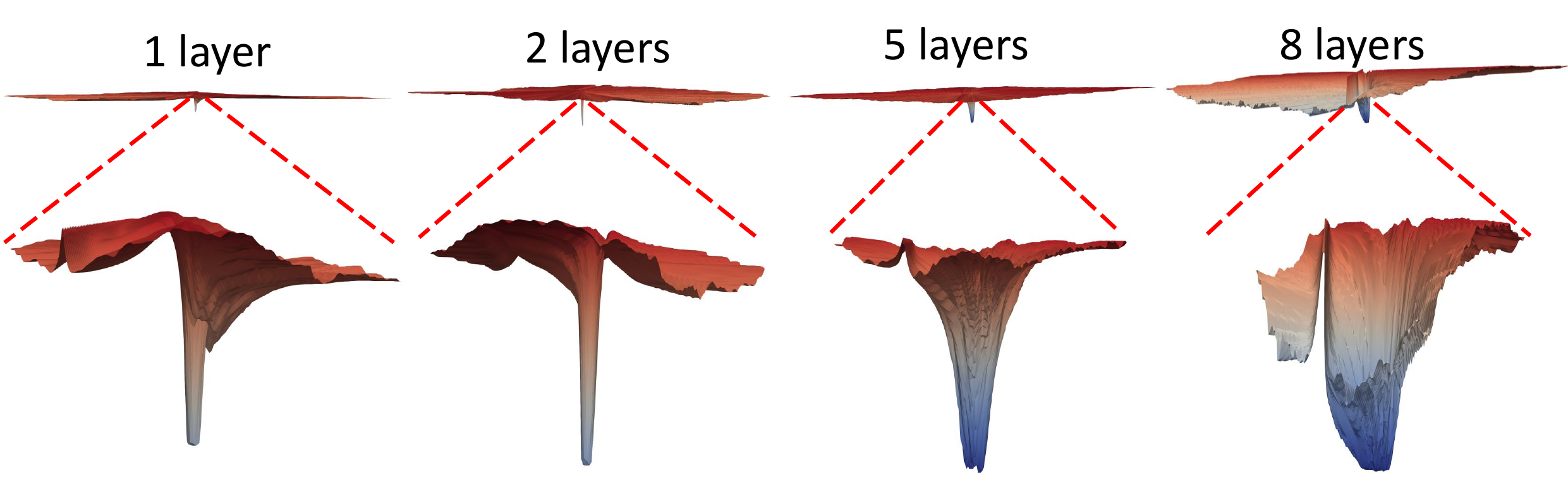}
        \caption{The landscape of instance Segmentation\_19. From left to right, the columns correspond to networks with \(K \in \{1,2,5,8\}\) layers. The first row is the landscape range from $[-10, +10]$ for both $\delta$ and $\eta$ direction. The second row is the landscape range from $[-1, +1]$ for both $\delta$ and $\eta$ direction.}
        \label{fig:land}
    \end{minipage}
\end{figure}

\begin{figure}[ht]
    \centering
    \includegraphics[width=.8\linewidth]{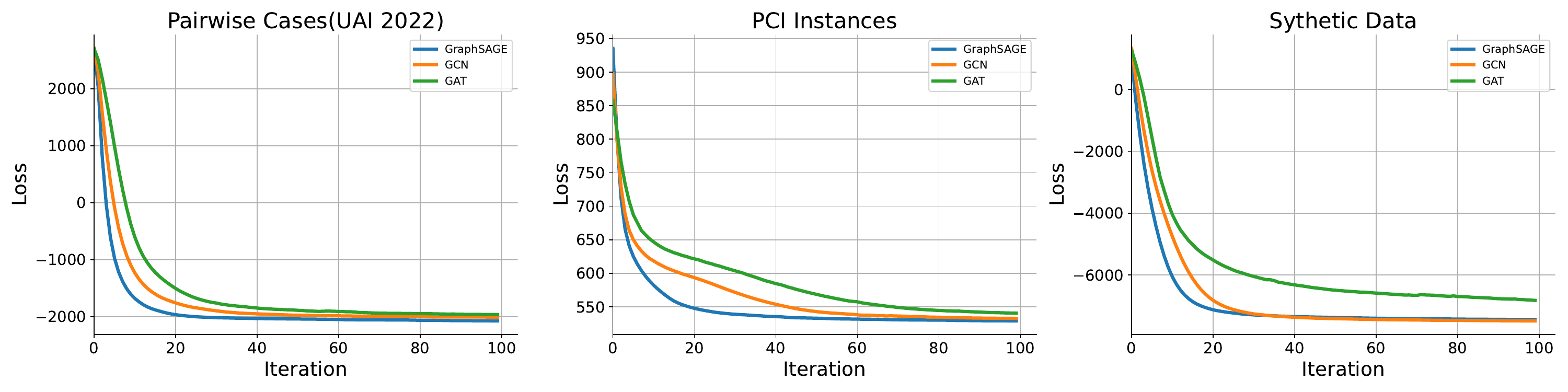}
    \caption{The average loss curves over UAI inference competition 2022 pairwise cases, PCI instances and synthetic instances using GraphSAGE, GCN and GAT as the GNN backbones.}
    \label{fig:GNN_backbone_curve}
\end{figure}

\begin{figure}[ht]
    \centering
    \includegraphics[width=0.4\textwidth]{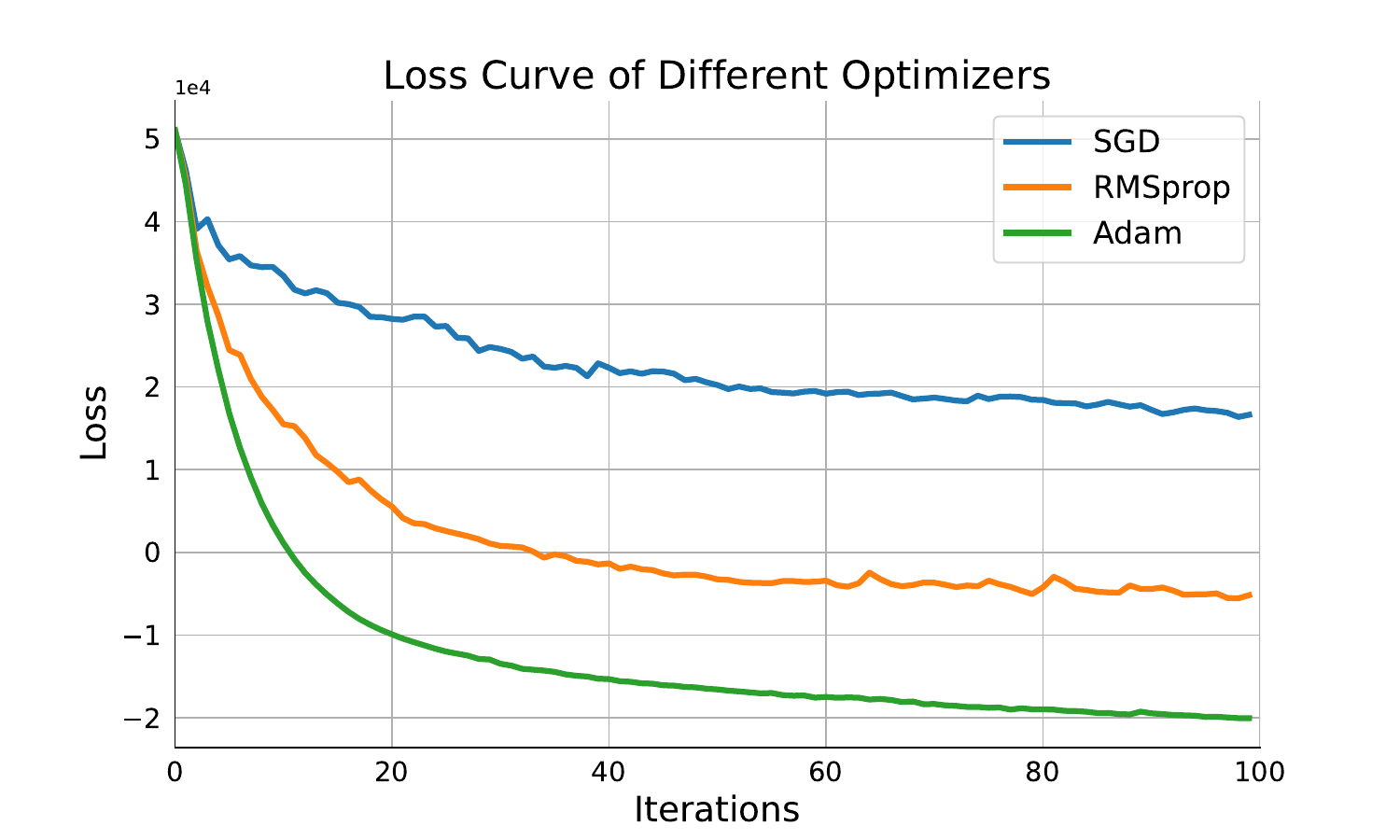}
    \caption{The average loss curves over UAI inference competition 2022 pairwise cases using different optimizers.}
    \label{fig:optimizer}
\end{figure}

\begin{figure}[ht]
        \centering
        \includegraphics[width=0.6\linewidth]{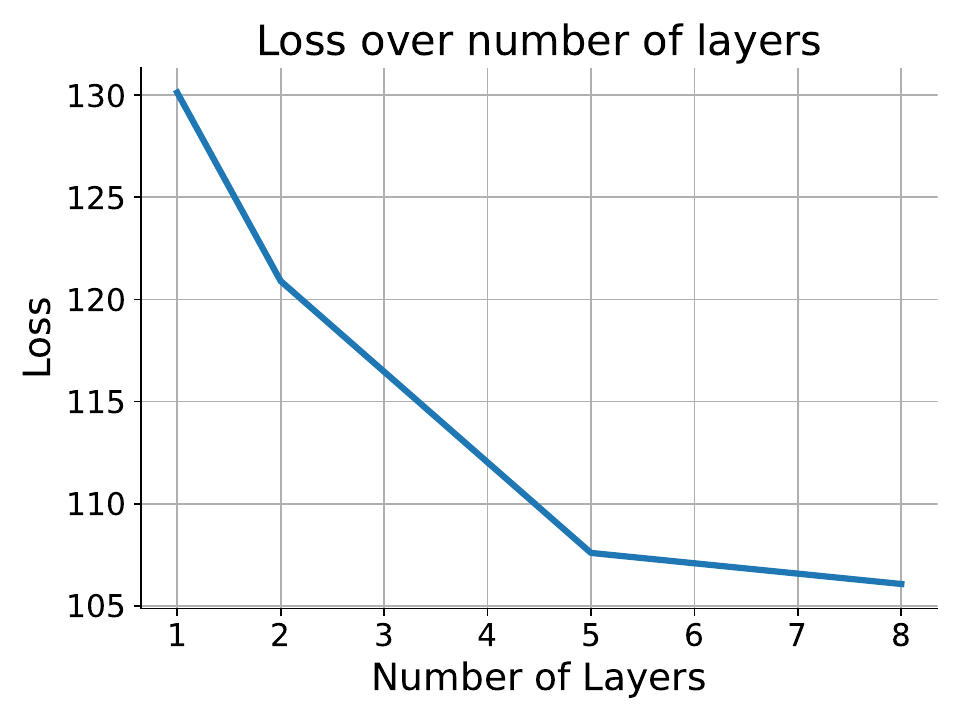}
        \caption{The training loss of instance Segmentation\_19 after convergence of using network layer number $\{1, 2, 5, 8\}$.}
        \label{fig:loss_layer}
\end{figure}

\begin{figure}[ht]
    \centering
    \includegraphics[width=0.55\linewidth]{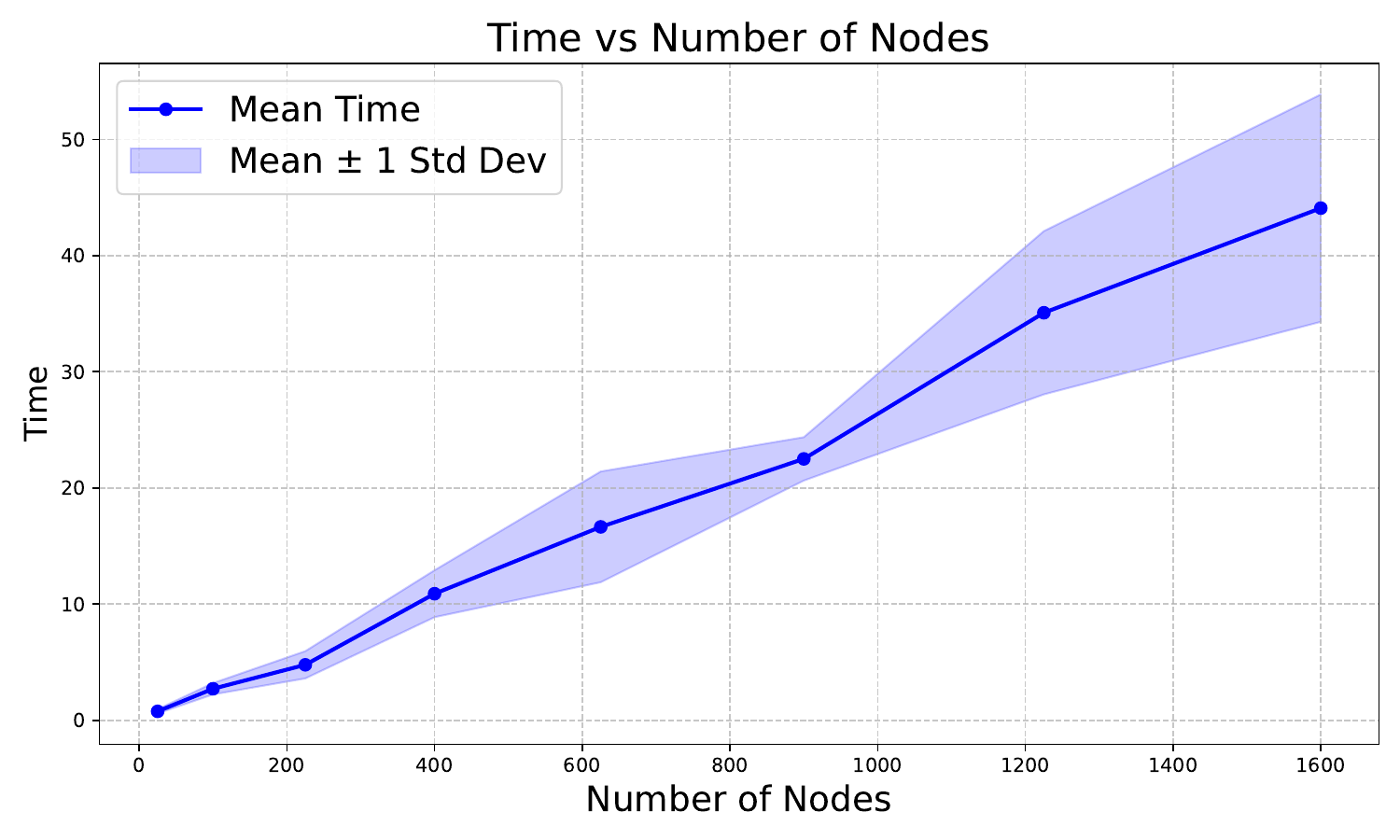}
    \caption{Wall-clock convergence time of \textsc{ReMAP} on grid graphs from $5^{2}$ to $50^{2}$ nodes. Each size is repeated five times. Convergence is a relative loss change below $10^{-4}$.}
    \label{fig:grid_time_vs_nodes}
\end{figure}

This appendix collects the controls pointed to from Section~\ref{sec:ablation}: GNN-parameter landscapes (Fig.~\ref{fig:land} and Fig.~\ref{fig:loss_layer}), backbone curves (Fig.~\ref{fig:GNN_backbone_curve}), optimizer curves (Fig.~\ref{fig:optimizer}), anytime PCI traces (Table~\ref{table:time}), and grid-graph wall-clock scaling (Fig.~\ref{fig:grid_time_vs_nodes}).

Fig.~\ref{fig:land} slices the GNN-parameter landscape of Segmentation\_19 at depths $K\in\{1,2,5,8\}$, using the perturbation setup of Appendix~\ref{app:vis_setup}. These plots live in GNN-parameter coordinates, not logit space. Theorem~\ref{thm:lossless} already implies that residual lifting does not enlarge the feasible logit set; depth only changes how $\rho_v(G)$ is parameterized and therefore the induced metric $I+JJ^{\top}$. Fig.~\ref{fig:loss_layer} shows the corresponding converged training loss. The two-layer four-hop locality result does not apply to the deeper experimental stacks.

Fig.~\ref{fig:GNN_backbone_curve} compares GraphSAGE, GCN, and GAT on UAI 2022 pairwise cases, PCI instances, and synthetic graphs ($1000$ nodes, average degree $4$ or $8$, random and Potts energies). GraphSAGE is the default backbone because neighbor aggregation is unweighted, matching MRF neighborhood structure (Table~\ref{table:GNNs} in Appendix~\ref{sec:GNNs}); the curves are consistent with that choice. Fig.~\ref{fig:optimizer} compares SGD, RMSprop, and Adam on UAI pairwise instances with learning rate $10^{-4}$, width $1024$, and eight layers. Adam is the practical optimizer used in the main tables.

Table~\ref{table:time} follows the UAI $1200$s protocol on the five real PCI instances, logging every $200$s. Toulbar2 solves PCI\_1--PCI\_3 within $200$s. On PCI\_4 and PCI\_5 it does not terminate within $1200$s and improves little after the first interval, while \textsc{ReMAP} continues to lower the incumbent. Timing for Toulbar2 on the hardest instances is coarse because the solver's log does not provide a dense trace.

Fig.~\ref{fig:grid_time_vs_nodes} measures wall-clock time to a relative loss change below $10^{-4}$ on grid graphs from $5^{2}$ to $50^{2}$ nodes, five repeats per size. The trend is approximately linear in the number of nodes and is consistent with the contraction-cost discussion in Appendix~\ref{app:complexity}.

\begin{table*}
    \begin{center}
        
    \caption{Time comparison between Toulbar2 and \textsc{ReMAP} on PCI instances. ``-" if the solving process is already stopped.}
    \label{table:time}
    \resizebox{\textwidth}{!}{
    \begin{tabular}{c|c|c|c|c|c|c|c}
        \toprule
         Instances & Algorithm/Solver  & 200s & 400s & 600s & 800s & 1000s & 1200s \\
         \midrule
         \multirow{2}{*}{PCI\_1} & Toulbar2  & 18.134 & - & - & - & - & - \\
                                & \textsc{ReMAP}   & 18.211 (10s in total) & - & - & - & - & - \\
         \midrule
         \multirow{2}{*}{PCI\_2} & Toulbar2  & 98.364 & - & - & - & - & - \\
                                & \textsc{ReMAP} & 98.446 (16s in total) & - & - & - & - & - \\
         \midrule
         \multirow{2}{*}{PCI\_3} & Toulbar2  & 1003.640  & - & - & - & - & - \\
                                & \textsc{ReMAP} & 1003.640  (71s in total) & - & - & - & - & - \\
         \midrule
         \multirow{2}{*}{PCI\_4} & Toulbar2  & 428.299 & 426.806 & 426.806 & 426.806 & 426.806 & 426.806 \\
                                & \textsc{ReMAP}& 408.508 & 407.6304  & 407.419 & - & - & - \\
         \midrule
         \multirow{2}{*}{PCI\_5} & Toulbar2  & 1128.244  & 1121.325 & 1121.325 & 1121.325 & 1121.325 & 1121.325 \\
                                & \textsc{ReMAP} & 1222.281 & 1086.899 & 1077.858 & 1074.3094 & 1070.8013 & 1069.875 \\
     \bottomrule
    \end{tabular}
    }
    \end{center}
\end{table*}

\section{Visualization setup}\label{app:vis_setup}
The core idea of the visualization technique proposed by \citep{visualloss} involves applying perturbations to the trained network parameters $\theta^*$ along two directional vectors, $\delta$ and $\eta$: $f(\alpha, \beta) = L(\theta^* + \alpha \delta + \beta \eta)$. By doing so, we can generate a 3-D representation of the landscape corresponding to the perturbed parameter space.

We sampled 250000 points in the $\alpha-\beta$ plane, where both $\alpha$ and $\beta$ range from -10 to 10, to obtain an overview of the loss function landscape.  Subsequently, we focused on the region around the parameter $\theta^*$ by sampling an additional 10,000 points in a narrower range, with $\alpha$ and $\beta$ both from $-1$ to $1$.

\section{Read UAI format files} \label{sec:read_uai}
An example data file in UAI format is provided in Box.~\ref{example:uai}. This Markov Random Field consists of 3 variables, each with 2 possible states. Detailed information can be found in the box, where we illustrate the meanings of different sections of the file. Notably, the potential tables are not normalized. During BP, these tables may be normalized to avoid numerical issues. For the energy transformation, however, we do not normalize them; each positive potential value $\phi$ is converted directly to $-\log\phi$.

\begin{tcolorbox}[title=Example.uai]
\footnotesize
MARKOV       \hspace{10pt} \textit{//Instance type}\\ 
3            \hspace{10pt} \textit{//Number of variables}\\
2 2 2        \hspace{10pt} \textit{//State number of each variable}\\
5            \hspace{10pt} \textit{//Number of cliques that has potentials}\\
1 0          \hspace{10pt} \textit{//1 means this clique is a variable, and the variable is 0.}\\
1 1\\
1 2\\
2 0 1          \hspace{10pt} \textit{//2 means this clique is an edge, the edge is (0, 1).}\\
3 0 1 2        \hspace{10pt} \textit{//3 means this clique includes 3 variables, and the clique is (0, 1, 2).}\\
\\
2           \hspace{10pt} \textit{//The number 2 indicates that the potential in the next line has two values.}\\
0.1 0.9     \hspace{10pt} \textit{//The potential of variable 0 is 0.1 for state 0 and 0.9 for state 1.}\\
\\
2 \\
0.1 10\\
\\
2\\
0.5 0.5\\
\\
4\\
0.1 1.0 1.0 0.1\textit{//The potential of the state combinations for variables 0 and 1 is given in the order of (0,0), (0,1), (1,0) and (1,1).}\\ 
\\
8\\
0.1 2.0 0.1 0.1 0.1 0.1 0.1 2.0 \hspace{10pt} \textit{//The potential of the state combinations for variables 0, 1, and 2 is given in the order of (0,0,0), (0,0,1), (0,1,0), (0,1,1), (1,0,0), and so on.}\\ 
\label{example:uai}
\end{tcolorbox}

Since the transformation of variable energies and clique energies follows the same procedure, we will use the edge $(0,1)$ to illustrate the transformation. Table~\ref{table:trans1} shows the unnormalized potentials for the edge $(0,1)$; Table~\ref{table:trans2} is obtained by applying $-\log(\cdot)$ entrywise to Table~\ref{table:trans1}.
 
\begin{table}[ht]
    \centering
    \begin{minipage}{0.49\textwidth}
        \centering
        \caption{$P(x_0, x_1)$}
        \label{table:trans1}

       \begin{tabular}{|c|c|c|}
        \hline
        \diagbox{$x_0$}{$x_1$}& 0 & 1 \\ 
                        \hline
                        0    &  0.1  & 1.0     \\
                        \hline
                        1   &  1.0  &  0.1     \\
        
        \hline
        
        \end{tabular}
    \end{minipage}
    \hfill 
    \begin{minipage}{0.49\textwidth}
        \centering
        \caption{$\theta_C(x_0, x_1)$ }
        \label{table:trans2}
        \begin{tabular}{|c|c|c|}
        \hline
        \diagbox{$x_0$}{$x_1$}& 0 & 1 \\ 
                        \hline
                            0    &  2.303  & 0     \\
                        \hline
                        1   &  0  &  2.303     \\
        
        \hline
        
        \end{tabular}
    \end{minipage}
\end{table}

\section{PCI problem formulation} \label{sec:trans_mip}
This appendix records the MIP-to-MRF conversion used for the PCI experiments in Section~\ref{sec:exp}. Table~\ref{table:pci_res} in the main text reports five real instances from a Chinese city. Table~\ref{table:pci_synthetic} below reports five larger synthetic PCI-style graphs generated from the same pairwise interference model, using the protocol of Section~\ref{sec:sys_exp} with $100$ iterations and a $3600$s Toulbar2 limit. On the synthetic family, LBP and TRBP remain far above the other two methods; Toulbar2 improves over message passing but is consistently above residual \textsc{ReMAP} as the graph grows.

The Mixed Integer Programming format of PCI problems is as follows: 

\begin{figure}[ht]
    \centering
    \begin{minipage}[t]{\textwidth} 
    {\footnotesize
        \begin{align} 
            \min \limits_{z, L} \quad & \sum \limits_{(i,j) \in \mathcal{E}} a_{ij} L_{ij}  \label{eq:pci} \\
        \mathrm{s.t.}\quad & z_{np} \in \{0, 1\} , \quad \forall n\in N, p \in P  \tag{\ref{eq:pci}{a}}\label{eq:pcia}\\
        & \sum \limits_{p\in P} z_{np} = 1 , \quad \forall n\in N    .  \tag{\ref{eq:pci}{b}}\label{eq:pcib} \\
        & \sum \limits_{p\in M_{ih}} z_{n_{i}p} + \sum \limits_{p\in M_{jh}} z_{n_jp} -1 \leq L_{ij}, \forall (i,j) \in \mathcal{E}, \forall h\in \{0, 1, 2\}. \tag{\ref{eq:pci}{c}}\label{eq:pcic} 
\end{align}
}
    \end{minipage}%
   
\end{figure}

where $n$ is the index for devices, and $N$ is the set of these indices. $P$ stands for the possible states of each device. $M_{ih}$ stands for the possible states set for node $n_i$. $L_{ij}$ is the cost when given a certain choices of the states of device $i$ and device $j$, $a_{ij}$ is the coefficient of the cost in the objective function. There is an $(i,j) \in \mathcal{E}$ means there exists interference between these two devices. 
 
When using MRF to model PCI problems, each random variable represents the identity state of the given node and the interference between devices would be captured by the pairwise energy functions. Next we will introduce how to transform the PCI problem from MIP form to MRF form.

In the original MIP formulation of the PCI problems, three types of constraints are defined. By combining Eq.~\ref{eq:pcia} and Eq.~\ref{eq:pcib}, we establish that each device must select exactly one state at any given time. Furthermore, the constraint in  Eq.~\ref{eq:pcic} indicates that interference occurs between two devices only when they select specific states. The overall impact on the system is governed by the value of $L_{ij}$ and its corresponding coefficient. Given that interference is always present, the objective is to minimize its extent.

To transform these problems into an MRF framework, we utilize Eq.~\ref{eq:pcib} to represent the nodes, where each instance of Eq.~\ref{eq:pcia} corresponds to the discrete states of a specific node. The constraints set forth in Eq.~\ref{eq:pcia} and Eq.~\ref{eq:pcib} ensure that only one state can be selected at any given time, thus satisfying those conditions automatically. By processing Eq.~\ref{eq:pcic}, we can identify the edges and their associated energies. If $z_{n_ip}$ and $z_{n_jp}$ appear in the same constraint from Eq.~\ref{eq:pcic}, we can formulate an edge $(i,j)$. By selecting different values for $z_{n_ip}$ and $z_{n_jp}$, we can determine the minimum value of $L_{ij}$ that maintains the validity of the constraint.

The product of $L_{ij}$ and $a_{ij}$ represents the energy associated with the edge $(i,j)$ under the combination of the respective states. Once the states of all nodes are fixed, the values of the edge costs also become fixed. This leads to the conclusion that the objective function is the summation of the energies across all edges. Since the PCI problems do not include unary terms, we can omit them during the transformation process. This establishes a clear pathway for converting the MIP formulation into an MRF representation, allowing us to leverage MRF methods for solving the PCI problems effectively.

\textbf{Example} \\
The original problem is 
 \begin{align}    
            \min \limits_{z, L} \quad & L_{1,2} + 3 L_{2,3} \notag \\
    \mathrm{s.t.} \quad & z_{np} \in \{0, 1\} , & \forall n\in \{1, 2, 3\}, p \in \{1, 2, 3\} \notag \\
    & \sum \limits_{p\in P} z_{np} = 1 , &\forall n\in \{1, 2, 3\}. \notag \\
    & z_{11} + z_{21} - 1 \leq L_{1,2} \notag\\
    & z_{13} + z_{22} - 1 \leq L_{1,2}\notag \\
    & z_{12} + z_{23} - 1 \leq L_{1,2} \notag\\
    & z_{21} + z_{31} - 1 \leq L_{2,3} \notag\\
    & z_{22} + z_{32} - 1 \leq L_{2,3} \notag\\
    & z_{23} + z_{33} - 1 \leq L_{2,3} \notag\\
\end{align}

Then the corresponding MRF problem is
\begin{equation}
    \min \theta_{1,2}(x_1, x_2) + \theta_{2,3}(x_2, x_3)
\end{equation}

the energy on edge $(x_1,x_2)$ and edge $(x_2, x_3)$  are shown in Table~\ref{table:energy1} and Table~\ref{table:energy2}.

\begin{table}
    \begin{minipage}{0.49\textwidth}
        \centering
        \caption{$E(x_1, x_2)$}
        \begin{tabular}{|c|c|c|c|}
            \hline
            \diagbox{$x_1$}{$x_2$} & $z_{21}$ & $z_{22}$ & $z_{23}$ \\ 
            \hline
            $z_{11}$ & 1 & 0 & 0 \\
            \hline
            $z_{12}$ & 0 & 0 & 1 \\
            \hline
            $z_{13}$ & 0 & 1 & 0 \\
            \hline
        \end{tabular} \label{table:energy1}
    \end{minipage}
    \hfill
    \begin{minipage}{0.49\textwidth}
        \centering
        \caption{$E(x_2, x_3)$}
        \begin{tabular}{|c|c|c|c|}
            \hline
            \diagbox{$x_2$}{$x_3$} & $z_{31}$ & $z_{32}$ & $z_{33}$ \\ 
            \hline
            $z_{21}$ & 3 & 0 & 0 \\
            \hline
            $z_{22}$ & 0 & 3 & 0 \\
            \hline
            $z_{23}$ & 0 & 0 & 3 \\
            \hline
        \end{tabular}\label{table:energy2}
    \end{minipage}
\end{table}

\begin{table}[ht]
\centering
\caption{Synthetic PCI instances, same pairwise MRF conversion as Table~\ref{table:pci_res}. Energy values; lower is better. Best in bold.}
\label{table:pci_synthetic}
\resizebox{\textwidth}{!}{
\begin{tabular}{cc|ccccc}
\toprule
\multicolumn{1}{c}{\multirow{2}{*}{Graph}} & \multicolumn{1}{c|}{\multirow{2}{*}{\#Nodes/\#Edges}} & \multicolumn{1}{c}{\multirow{2}{*}{LBP}} & \multicolumn{1}{c}{\multirow{2}{*}{TRBP}} & \multicolumn{1}{c}{\multirow{2}{*}{Toulbar2}} & \multicolumn{2}{c}{\textsc{ReMAP}} \\
\cline{6-7}
& & & & & Energy & Loss \\
\midrule
PCI\_synthetic\_1 & 280/9678 & 564198.000 & 568082.000 & 522857.923 & $\mathbf{496015.5\pm 6307.363}$ & $496013.662\pm 6297.169$ \\
PCI\_synthetic\_2 & 526/34500 & 2.092e+06 & 2.084e+06 & 2.064e+06 & $\mathbf{1.923e+06\pm 9977.015}$ & $1.923e+06\pm 10007.739$ \\
PCI\_synthetic\_3 & 1000/49950 & 2.932e+06 & 2.908e+06 & 2.856e+06 & $\mathbf{2.665e+06\pm 4555.868}$ & $2.664e+06\pm 4468.965$ \\
PCI\_synthetic\_4 & 1500/78770 & 4.568e+06 & 4.532e+06 & 4.534e+06 & $\mathbf{4.215e+06\pm 13500.602}$ & $4.214e+06\pm 13252.456$ \\
PCI\_synthetic\_5 & 2000/120024 & 6.807e+06 & 6.904e+06 & 7.023e+06 & $\mathbf{6.542e+06\pm 19789.758}$ & $6.540e+06\pm 19782.638$ \\
\bottomrule
\end{tabular}
}
\end{table}

\section{Limitations} \label{app:limitation}
Preconditioning is not global optimality. The residual bypass makes the lifting lossless and guarantees at least the instantaneous free-logit dissipation. It does not make a nonconvex trajectory reach a global solution from every initialization. The local rate requires entry into a MAP-margin neighborhood, and the factor $(1+\beta)^{-1}$ requires residual alignment $\|J_v^{\top}q\|_{2}^{2}\ge\beta\|q\|_{2}^{2}$; $\beta$ is trajectory-dependent and is not implied by the architecture.

High order can be exponentially large. For explicit dense tables, time and memory grow with $\prod_{i\in C}s_i$. The method scales to many nodes only when the total factor representation and contraction fit memory. Structured kernels or low-rank factors are necessary for genuinely large order.

The PCI parameterization ablation compares terminal discrete energies of residual, GNN-only, and free logits. It does not estimate $\beta_t=\|J_v^{\top}q\|_{2}^{2}/\|q\|_{2}^{2}$ along the residual trajectory, and it is not a matched-compute anytime test.

Heuristic optimization. Adam, restarts, and optional temperature schedules have no global convergence guarantee here. Conditional rounding guarantees a discrete energy no worse than the current relaxed value, but neither quantity is guaranteed close to $E_{\star}$ without an external lower bound.

Best use as a portfolio component. The most credible operational role is to generate GPU incumbents for difficult instances and pass them to an exact or dual solver, yielding the certified gap $E(x)-B$.

\end{document}